\documentclass[10pt,twocolumn,letterpaper]{article}

\usepackage{cvpr}      

\usepackage{graphicx}
\usepackage{amsmath}
\usepackage{amssymb}
\usepackage{multicol, blindtext}
\usepackage{subcaption}
\usepackage{cite}
\usepackage{soul}
\usepackage{units}
\usepackage{array}
\usepackage{multirow}
\usepackage{caption}
\usepackage{float}
\usepackage{booktabs}
\usepackage{times}
\usepackage{tabularx}
\usepackage{nicefrac}
\usepackage{enumitem}

\usepackage{adjustbox}

\newcolumntype{R}[2]{
    >{\adjustbox{angle=#1,lap=\width-(#2)}\bgroup}%
    l
    <{\egroup}%
}

\newcolumntype{L}[1]{>{\raggedright\arraybackslash}p{#1}}
\newcolumntype{C}[1]{>{\centering\arraybackslash}p{#1}}
\newcolumntype{R}[1]{>{\raggedleft\arraybackslash}p{#1}}
\usepackage{pifont}
\newcommand{\cmark}{\ding{51}}
\newcommand{\xmark}{\ding{55}}

\usepackage{caption}
\usepackage{xcolor}
\usepackage{mathtools}
\usepackage{xfrac}
\usepackage{wasysym}
\usepackage{algorithm}%
\usepackage{algpseudocode}%
\usepackage[accsupp]{axessibility}  

\makeatletter

\@namedef{ver@everyshi.sty}{}
\newcommand\notsotiny{\@setfontsize\notsotiny{6.31415}{7.1828}} 

\makeatother

\usepackage{comment}
\usepackage{pgfplots}
\usepackage{tikz} 
\usetikzlibrary{patterns}
\pgfplotsset{compat=newest}

\usepackage[pagebackref,breaklinks,colorlinks]{hyperref}

\usepackage[capitalize]{cleveref}
\crefname{section}{Sec.}{Secs.}
\Crefname{section}{Section}{Sections}
\Crefname{table}{Table}{Tables}
\crefname{table}{Tab.}{Tabs.}

\newcommand{\probP}{\kern0.15em \text{I\kern-0.15em P}}

\newcommand{\wet}{\textvisiblespace \kern0.25em (ours)}

\newcommand{\both}{\textvisiblespace \kern-0.45em * (ours)}

\definecolor{water_color}{RGB}{69,156,238}
\definecolor{eth_orange}{RGB}{255,126,40}
\definecolor{eth_green}{RGB}{153,204,51}
\definecolor{eth_blue}{RGB}{169,204,242}
\definecolor{eth_red}{RGB}{230,140,132}
\definecolor{eth_gray}{RGB}{126,126,126}

\newcommand{\PAR}[1]{\vspace{-0.2eM}\vskip4pt \noindent{\bf #1}}

\def\cvprPaperID{6729} 
\def\confName{CVPR}
\def\confYear{2023}

\begin{document}

\title{Gated Stereo: \\Joint Depth Estimation from Gated and Wide-Baseline Active Stereo Cues\vspace{-1eM}
}
\author{\centerline{\hspace{-0.5eM}Stefanie Walz$^1$ Mario Bijelic$^2$ Andrea Ramazzina$^1$ Amanpreet Walia$^3$ Fahim Mannan$^3$
Felix Heide$^{2,3}$}
\and
\centerline{$^1$Mercedes-Benz \quad $^2$Princeton University \quad $^3$Algolux}}
\maketitle

\vspace{-1eM}

\begin{abstract}
\vspace{-0.7eM}
We propose Gated Stereo, a high-resolution and long-range depth estimation technique that operates on active gated stereo images. Using active and high dynamic range passive captures, Gated Stereo exploits multi-view cues alongside time-of-flight intensity cues from active gating. 
To this end, we propose a depth estimation method with a monocular and stereo depth prediction branch which are combined in a final fusion stage. Each block is supervised through a combination of supervised and gated self-supervision losses. To facilitate training and validation, we acquire a long-range synchronized gated stereo dataset for automotive scenarios. We find that the method achieves an improvement of more than \unit[50]{\%} MAE compared to the next best RGB stereo method, and \unit[74]{\%} MAE to existing monocular gated methods for distances up to \unit[160]{m}. 
Our code, models and datasets are available \href{https://light.princeton.edu/gatedstereo/}{here}\footnote{\scriptsize\url{https://light.princeton.edu/gatedstereo/}\label{link}}.
\end{abstract}

\section{Introduction}

Long-range high-resolution depth estimation is critical for autonomous drones, robotics, and driver assistance systems. 
Most existing fully autonomous vehicles strongly rely on scanning LiDAR for depth estimation~\cite{schwarz2010lidar,WaymoDataset}. While these sensors are effective for obstacle avoidance the measurements are often not as semantically rich as RGB images. LiDAR sensing also has to make trade-offs due to physical limitations, especially beyond 100 meters range, including range range versus eye-safety and spatial resolution. Although recent advances in LiDAR sensors such as, MEMS scanning~\cite{mems:mirrors:lidar:review} and photodiode technology~\cite{villa2012spad} have drastically reduced the cost and led to a number of sensor designs with  $\approx$ $100$ - $200$ scanlines, these are still significantly lower resolutions than modern HDR megapixel camera sensors with a vertical resolution more than $\approx$ $5000$ pixels. 
However, extracting depth from RGB images with monocular methods is challenging as existing estimation methods suffer from a fundamental scale ambiguity~\cite{eigen2014depth}. Stereo-based depth estimation methods resolve this issue but need to be well calibrated and often fail on texture-less regions and in low-light scenarios when no reliable features, and hence triangulation candidate, can be found. 

To overcome the limitations of existing scanning LiDAR and RGB stereo depth estimation methods, a body of work has explored gated imaging~\cite{heckman1967,grauer2014active,Bijelic2018,Busck2004, Busck2005, Andersson2006}. Gated imagers integrate the transient response from flash-illuminated scenes in broad temporal bins, see Section~\ref{sec:GatedStereoImaging} for more details. 
This imaging technique is robust to low-light, and adverse weather conditions \cite{Bijelic2018} and the embedded time-of-flight information can be decoded as depth. Specifically, Gated2Depth \cite{gated2depth2019} estimates depth from three gated slices and learns the prediction through a combination of simulation and LiDAR supervision. Building on these findings, recently, Walia et al.~\cite{gated2gated} proposed a self-supervised training approach predicting higher-quality depth maps. However, both methods have in common that they often fail in conditions where the signal-to-noise ratio is low, e.g., in the case of strong ambient light. 

We propose a depth estimation method from gated stereo observations that exploits both multi-view and time-of-flight cues to estimate high-resolution depth maps. We propose a depth reconstruction network that consists of a monocular depth network per gated camera and a stereo network that utilizes both active and passive slices from the gated stereo pair. The monocular network exploits depth-dependent gated intensity cues to estimate depth in monocular and low-light regions while the stereo network relies on active stereo cues. Both network branches are fused in a learned fusion block. Using passive slices allows us to perform robustly under bright daylight where active cues have a low signal-to-noise ratio due to ambient illumination. To train our network, we rely on supervised and self-supervised losses tailored to the stereo-gated setup, including ambient-aware and illuminator-aware consistency along with multi-camera consistency. To capture training data and assess the method, we built a custom prototype vehicle and captured a stereo-gated dataset under different lighting conditions and automotive driving scenarios in urban, suburban and highway environments across 1000~km of driving. 

Specifically, we make the following contributions:\vspace{-0.2eM}
\begin{itemize}
\itemsep-0.3em
  \item We propose a novel depth estimation approach using gated stereo images that generates high-resolution dense depth maps from multi-view and time-of-flight depth cues. 
  \item We introduce a depth estimation network with two different branches for depth estimation, a monocular branch and a stereo branch, that use active and passive measurement, and a semi-supervised training scheme to train the estimator. 
  \item We built a prototype vehicle to capture test and training data, allowing us to assess the method in long-range automotive scenes, where we reduce the MAE error by \unit[50]{\%} to the next best RGB stereo method and by \unit[74]{\%} on existing monocular gated methods for distances up to \unit[160]{m}. 
\end{itemize}

\section{Related Work}

\PAR{Depth from Time-of-Flight.}
Time-of-Flight (ToF) sensors acquire depth by estimating the round travel time of light emitted into a scene and returned to the detector. Broadly adopted approaches to time-of-flight sensing include correlation time of flight cameras \cite{hansard2012time, kolb2010time, lange00tof}, pulsed ToF sensors \cite{schwarz2010lidar} and gated illumination with wide depth measuring bins \cite{heckman1967,grauer2014active}.
Correlation time-of-flight sensors\cite{hansard2012time, kolb2010time,
lange00tof} flood-illuminate a scene and estimate the depth from the phase difference of the emitted and received light. This allows for precise depth estimation with high spatial resolution but due to its sensitivity to ambient light existing correlation time-of-flight detectors have been limited to indoor applications. In contrast, pulsed light ToF systems \cite{schwarz2010lidar} measure the round trip time directly from a single light pulse emitted to a single point in the scene. Although a single point measurement offers high depth precision and signal-to-noise ratio, this acquisition process mandates scanning to allow long outdoor distances and, as such, drastically reduces spatial resolution in dynamic scenes. In addition, pulsed LiDAR measurements can drastically degrade in adverse weather~\cite{BenchmarkLidar,LIBRE,Jokela} due to backscattered light from fog or snow. Gated cameras~\cite{heckman1967,grauer2014active,Bijelic2018} accumulate flood-illuminated light over short temporal bins limiting the visible scene to certain depth ranges. As a result, gated cameras gate-out backscatter and at short-range~\cite{Bijelic2018} and reconstruct coarse depth~\cite{Busck2004, Busck2005, Andersson2006}. 

\PAR{Depth Estimation from Monocular and Stereo Intensity Images.}
Depth estimation from single \cite{MovingPeopleMovingCameras, godard2017unsupervised,li2022depthformer,guizilini20203d}, single images with sparse LiDAR points \cite{tang2019sparse2dense, wong2021unsupervised, park2020nonRGBLidar, hu2020PENetRGBLidar, guidenetRGBLidar},  stereo image pairs \cite{chang2018pyramid,TransformerStereo,badki2020Bi3D,yang2019hsm} or stereo with sparse LiDAR \cite{VolPropagationNetStereoLidar, SLFNetStereoLidar} is explored in a large body of work. 
Monocular depth imaging approaches \cite{MovingPeopleMovingCameras} offer low cost when a single CMOS camera is used, reduced footprint, especially compared to LiDAR systems, and, hence, also can be applied across application domains. However, monocular depth estimation methods inherit a fundamental scale ambiguity problem that can only be resolved by vehicle speed or LiDAR ground-truth depth measurements at test-time \cite{guizilini20203d}. Stereo approaches, on the other hand, allow triangulating between two different views resolving the scale ambiguity \cite{chang2018pyramid}. As a result, these methods allow for accurate long-range depth prediction when active sensors are not present. To learn the depth prediction from stereo intensity images, existing methods employ supervised \cite{chang2018pyramid,li2022depthformer, eigen2014depth,chang2018pyramid,jaritz2018sparse,ma2018sparse,MovingPeopleMovingCameras,Mayer2016,Kendall2017} and unsupervised learning techniques \cite{Zhou2017, godard2019digging,guizilini20203d, Garg2016, godard2017unsupervised}. 
Supervised stereo techniques often rely on time-of-flight data \cite{eigen2014depth, chang2018pyramid,jaritz2018sparse,ma2018sparse} or multi-view data \cite{MovingPeopleMovingCameras,Mayer2016,Kendall2017} for supervision. As a result, the collection of suitable dense ground-truth data can be challenging. Specifically, existing work \cite{geiger2012we,Uhrig2017THREEDV} aims to compensate for the sparsity of LiDAR ground-truth measurements through ego-motion correction and acquisition of multiple point clouds. Moreover, such aggregated LiDAR ``ground-truth'' depth is incorrect in scattering media \cite{Bijelic_2020_STF}.
To tackle this challenge and exploit large datasets of video data without ground-truth LiDAR depth present, self-supervised stereo approaches exploit multiview geometry by aligning stereo image pairs~\cite{Garg2016, godard2017unsupervised} or they make use of image view synthesis between temporally consecutive 
frames~\cite{Zhou2017, godard2019digging,guizilini20203d}.  
Garg et al.~\cite{Garg2016} train a network to predict disparities from monocular camera images by encouraging consistency when warped to stereo images. Followup work~\cite{Ummenhofer2017,godard2017unsupervised} extends this idea to warp temporally consecutive stereo captures. To perform the warping correctly for these methods, two networks are necessary one predicting the depth and a second one predicting a rigid body transformation between two temporally adjacent frames. 
Existing work on depth prediction has investigated diverse neural architectures for depth estimation networks~\cite{Garg2016,godard2019digging,guizilini20203d,TransformerStereo,badki2020Bi3D,yang2019hsm,li2022depthformer} and extensions in the loss formulation \cite{godard2017unsupervised,vijayanarasimhan2017sfm, yin2018geonet, ranjan2019competitive, godard2019digging, luo2019every, dai2020self, guizilini20203d}. Recently, RAFT-Stereo \cite{lipson2021raft} relies on iterative refinement over the cost volume at high resolution, thanks to the construction of a lighter cost volume and the employment of 2D convolution instead of 3D convolutions, which are memory and computationally intensive. All depth estimation methods discussed above, which are based on passive imaging, can fail in low-light or low-contrast scenarios that active gated methods~\cite{gated2depth2019} tackle using illumination. Alternate approaches employ sparse LiDAR measurements \cite{tang2019sparse2dense, wong2021unsupervised, park2020nonRGBLidar, hu2020PENetRGBLidar, guidenetRGBLidar, VolPropagationNetStereoLidar, SLFNetStereoLidar} not only for supervised training but also during inference time to overcome the scale ambiguity from monocular approaches, but they come with the drawback that temporal LiDAR distortions and scan pattern artefacts are passed through.

\PAR{Depth Estimation from Gated Images.}
Gated depth estimation methods with analytical solutions guiding the depth estimation \cite{Laurenzis2007, Laurenzis2009, Xinwei2013} had been first proposed over a decade ago. Recently, learned Bayesian approaches \cite{adam2017bayesian,schober2017dynamic} and approaches employing deep neural networks \cite{gated2depth2019, gated2gated} have achieved dense depth estimation at long-range outdoor scenarios and in low-light environments. All of these existing methods rely on monocular gated imaging systems, which are able to deliver similar performances to passive color stereo approaches \cite{gated2depth2019,gated2gated}. Gruber et al.~\cite{gated2depth2019} introduce a fully supervised depth prediction network leveraging pretraining on fully synthetic data performing on par with traditional stereo approaches. Recently Walia et al.~\cite{gated2gated} proposed a self-supervised gated depth estimation method. Although their approach resolves the scale ambiguity, it still suffers in bright daylight in the absence of depth cue, and at long ranges due to depth quantization and lack of relative motion during training.
In this work, we tackle these issues with a wide-baseline stereo-gated camera to estimate accurate depth in all illumination conditions and at long ranges.

\begin{figure}[t!]
    \centering
    \includegraphics[width=0.49\textwidth]{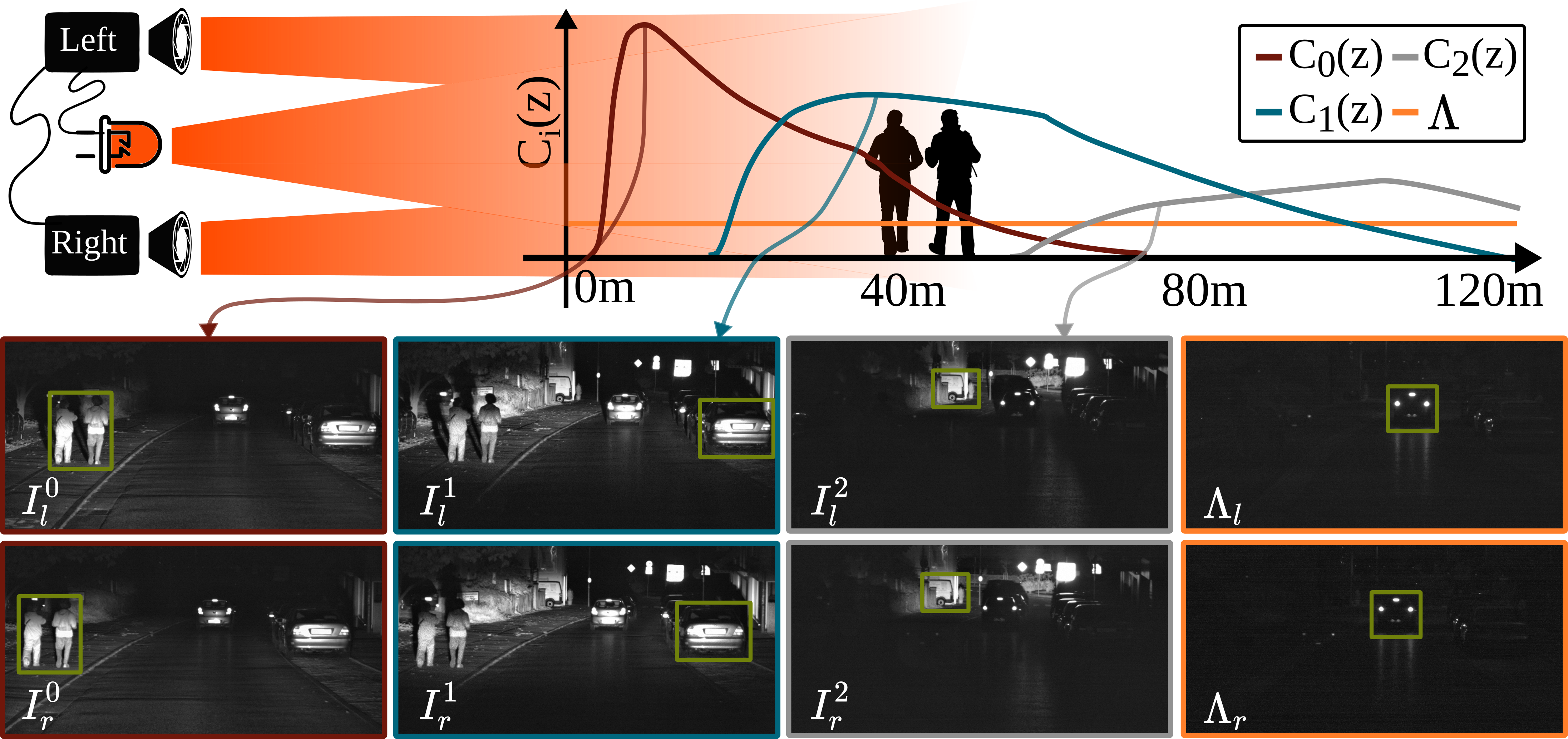}
    \caption{The proposed stereo gated camera consists of two gated cameras and a single flood-lit pulsed illumination source. Varying the delay between illumination and the synchronized cameras results in different range-intensity profiles $C_{k}$ describing the pixel-intensity for distance $z$ for each camera in addition to disparity $d$. For image formation in bright airlight, an additional passive component $\Lambda$ is required. The resulting images for left and right camera positions illustrating gating and parallax in an example scene are illustrated at the bottom. } 
    \label{fig:rip}
    \vspace{-3mm}
\end{figure}

\begin{figure*}[!ht]
\vspace{-6mm}
    \centering
    \includegraphics[width=1\textwidth]{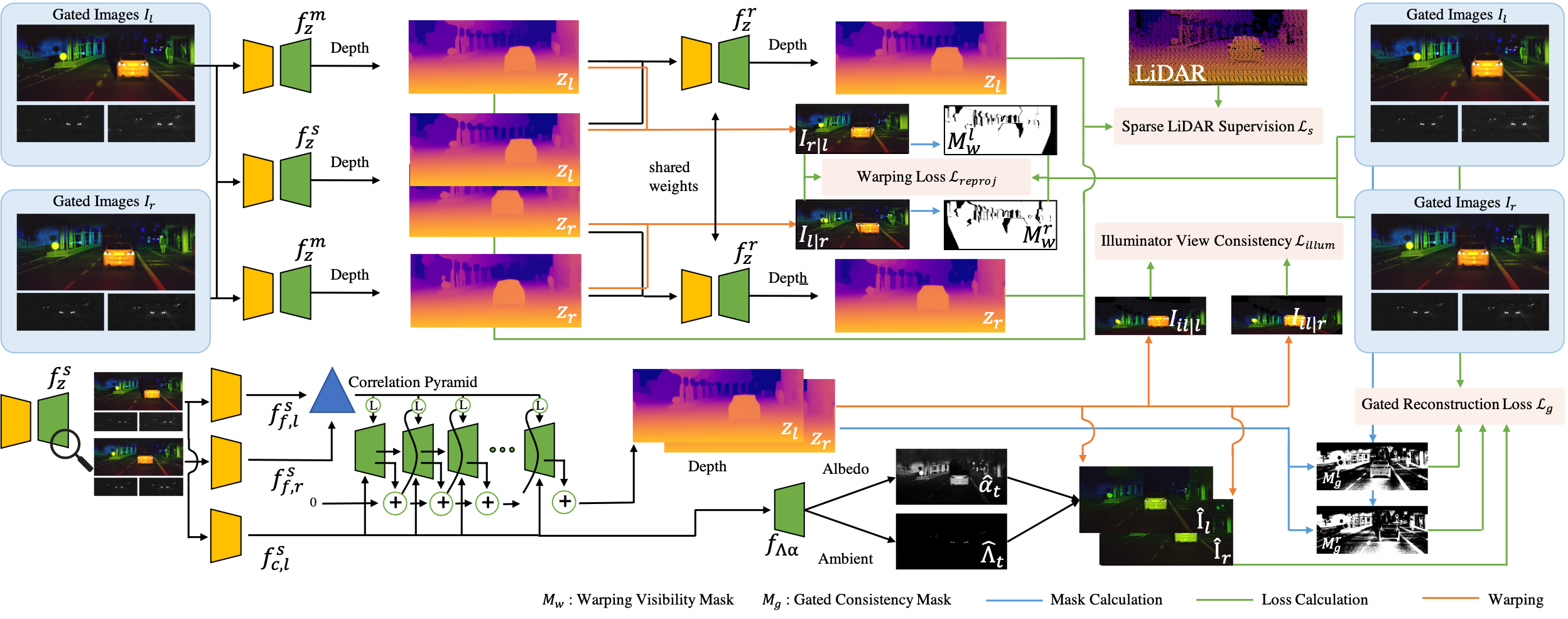}
    \vspace{-2eM}
    \caption{The proposed model architecture is composed of a stereo ($f^s_z$), two monocular ($f^m_z$), and two fusion ($f^r_z$) networks with shared weight. The fusion network combines the output of the monocular and stereo networks to obtain the final depth image for each view. Both stereo and monocular networks use active and passive slices as input, with the stereo network using the passive slices as context and includes a decoder ($f_{\Lambda\alpha}$) for albedo and ambient estimation which are used for gated reconstruction. The loss terms are applied to the appropriate pixels using masks that are estimated from the inputs and outputs of the networks.\vspace{-3mm}} 
    \label{fig:arch}
    
\end{figure*}
\vspace{-1.0mm}
\section{Gated Stereo Imaging}\label{sec:GatedStereoImaging}

\vspace{-0.25eM}
This section introduces the proposed gated stereo camera. 
We propose a synchronized gated camera setup with a wide baseline of $b= $\unit[0.76]{m}. After flood-illuminating the scene with a single illuminator, we capture three synchronized gated and passive slices with two gated cameras. Synchronizing two gated cameras requires not only the trigger of individual single exposures as for traditional stereo cameras, but the transfer of gate information for each slice with nano-second accuracy. This level of synchronization allows us to extract slices with gated multi-view cues. 

Specifically, after the emission of a laser pulse $p$ at time $t=0$, the reflection of the scene gets integrated on both camera sensors after a predefined time delay $\xi$ identical on both cameras. Only photons arriving in a given temporal gate are captured with the gate function $g$ allowing to integrate implicit depth information into 2D images. Following Gruber \textit{et al.}~\cite{gruber2018learning}, the distance-dependent pixel intensities are described by so-called range-intensity-profiles $C_k(z)$ which are independent of the scene and given by, 
\begin{equation}
\begin{aligned}
I^{k}(z,t)\;&=\;\alpha\,C_{k}(z,t),\\
    \;&=\;\alpha \int\limits_{-\infty}^{\infty} g_k(t-\xi)p_k\left(t\,-\,\cfrac{2z}{c}\right)\beta(z)dt,
\label{eq:gated_img}
\end{aligned}
\end{equation}
where $I^{k}(z,t)$ is the gated exposure, indexed by $k$ for the slice index at distance $z$ and time $t$; $\alpha$ is the surface reflectance (albedo), and $\beta$ the attenuation along a given path due to atmospheric effects. Both image stacks are rectified and calibrated such that epipolar lines in both cameras are aligned along the image width and disparities $d$ can be estimated. Epipolar disparity is consistent with the distance $z=\frac{bf}{d}$, where $f$ is the focal length, providing a depth cue across all modulated and unmodulated slices.

In the presence of ambient light or other light sources as sunlight or vehicle headlamps, unmodulated photons are acquired as a constant $\Lambda$ and added to the Eq.~\ref{eq:gated_img},
\begin{equation}
I^{k}(z)\;=\;\alpha\,C_{k}(z)+\,\Lambda.
\label{eq:final_gated_eq}
\end{equation}
Independently from ambient light, a dark current $D^k_v$ depending on the gate settings is added to the intensity count,
\begin{equation}
I^{k}_v(z)\;=\;\alpha\,C_{k}(z)+\,\Lambda + D^k_v,
\label{eq:final_gated_eq_dark_current}
\end{equation}
which we calibrate for each gate $k$ and camera $v$.
We adopt the Poisson-Gaussian noise model from ~\cite{gated2gated}. In contrast to prior work~\cite{gated2depth2019,gated2gated}, we also capture two unmodulated passive exposures in an HDR acquisition scheme. So specifically, we use three gated exposures $C_1, C_2, C_3$ with the same profile as in~\cite{gated2depth2019} and two additional passive images without illumination, that is, $C_4 = C_5 = 0$, and HDR-like fixed exposure times of \unit[21]{\textmu s} and \unit[108]{\textmu s} at daytime and \unit[805]{\textmu s} and \unit[1745]{\textmu s} at night time. This allows us to recover depth simultaneously from stereo-gated slices and passive stereo intensity cues with the same camera setup. The proposed system captures these images at 120~Hz, natively, allowing for a per-frame update of 24~Hz, which is about 2$\times$ the update rate of recent commercial scanning LiDAR systems, e.g., Luminar Hydra or Velodyne Alpha Puck.

\section{Depth from Gated Stereo}
\vspace{-1mm}

In this section, we propose a depth estimation method that exploits active and passive multi-view cues from gated images. Specifically, we introduce a joint stereo and monocular network that we semi-supervise this network using several consistency losses tailored to gated stereo data. 
In the following, we first describe the proposed network architecture before describing the semi-supervision scheme.
\subsection{Joint Stereo-Mono Depth Network}
The proposed depth estimation network is illustrated in Fig.~\ref{fig:arch}, which has a stereo and monocular branches, and a final fusion network that combines the outputs from these branches to produce the final depth map.

\PAR{Monocular Branch.}
The monocular network, $f^m_z:I \rightarrow z^m$, estimates absolute depth for a single gated image $I$ from either of the two imagers.
Unlike monocular RGB images, monocular gated images encode depth-dependent intensities which can be used by monocular depth networks to estimate \emph{scale-accurate} depth maps \cite{gated2depth2019, gated2gated}. The proposed monocular gated network uses a DPT \cite{ranftl2021vision}-type architecture and outputs inverse depth bounded in $\left[0, 1\right]$ which results in absolute depth between $\left[1, \infty\right]$. For network details, we refer to the Supplemental Material.
\PAR{Stereo Branch.}
The stereo branch, $f^s_z:(I_l,I_r)\rightarrow(z_l^s, z_r^s)$, estimates disparity from a pair of stereo images and outputs the depth for the left and right images $z_l$ and $z_r$ respectively. The network architecture is based on RAFT-Stereo \cite{lipson2021raft} with all three active gated slices and two passive captures concatenated to a 5-dimensional input. The feature extractor is replaced with HRFormer~\cite{YuanFHLZCW21}, which is able to extract robust high-resolution features for downstream stereo matching. The left and right slice features  $f^s_{f,l}$ and $f^s_{f,r}$ are given as input to the correlation pyramid module and the context feature $f^s_{c,l}$ are used as input for the GRU layers (see Fig.~\ref{fig:arch} bottom-left). Furthermore, the context features are fed to a decoder, $f_{\Lambda\alpha}$, to estimate the albedo and ambient components for gated slice reconstruction.

\PAR{Stereo-Mono Fusion.}
Monocular gated depth estimates suffer from depth quantization due to the depth binning of gated slices, failure in the presence of strong ambient illumination, and illuminator occlusion. 
Stereo methods, in isolation, suffer from inherent ambiguity in partially occluded regions and can fail when one of the views is completely obstructed, e.g., by lens occlusions and bright illumination.
Previous work \cite{chen2021revealing} proposed distilling the monocular network with the stereo output, and distilling the stereo network with fused pseudo-labels.
Departing from that approach, we use a light-weight 4-layer ResUNet \cite{zhang2018road} network, $f^r_z:(z^m, z^s, I)\rightarrow z^f$, that takes in monocular and stereo depth with the corresponding active and passive slices as input and produce a single fused depth map as output. The active and passive slices provide additional cues for the fusion network.

\vspace{5pt}
\noindent
With the proposed depth estimation network in hand, we propose a set of stereo and monocular semi-supervised training signals for actively illuminated gated stereo pairs along with high dynamic passive captures.

\subsection{Depth and Photometric Consistency}
\label{sec:consistencylosses}

We rely on self-supervised consistency losses and sparse supervised losses as following.

\begin{figure}[!t]
    \centering
    \vspace{-0.1eM} 
    \includegraphics[width=0.49\textwidth]{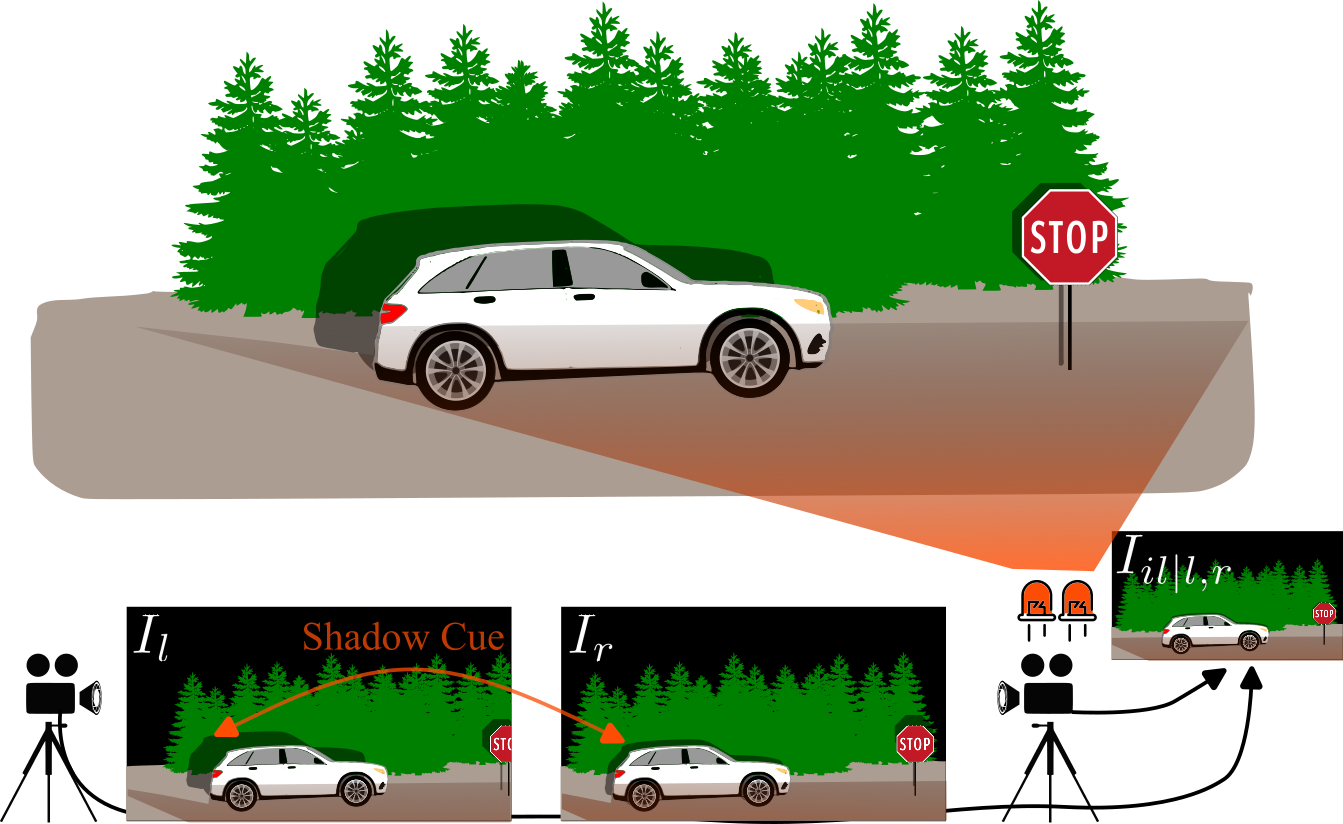}   
    \vspace{-3mm}
    \caption{Scene regions occluded in the illuminator view will be in shadow in the two views (left, middle), and shadowless after projecting to the illuminator viewpoint (right).}
    \label{fig:shadowloss}
    \vspace{-3mm}
\end{figure}

\PAR{Left-Right Reprojection Consistency.}
This loss enforces the photometric consistency between the left and right gated images given the per-pixel disparity,
\begin{equation}
\begin{aligned}
\mathcal{L}_{reproj} &=  \mathcal{L}_p(\mathcal{M}^{o}_{l|r} \odot I_l, \mathcal{M}^{o}_{l|r} \odot I_{l|r}),
\label{eq:reproj_loss}
\end{aligned}
\end{equation}
with $I_{l|r}$ the left image warped into the right view using the predicted disparity $d_{l}$. Here, $\mathcal{L}_p$ \cite{godard2017unsupervised} is a similarity loss based on the structural similarity (SSIM) metric \cite{ssim} and the $L_1$ norm, $\mathcal{L}_p(a,b)=0.85\frac{1 - SSIM(a, b)}{2} + 0.15\|a - b\|_1$. The occlusion mask $\mathcal{M}^{o}_{l|r}$ indicates pixels in the left image that are occluded in the right image and is defined as a soft mask for better gradient flow, $\mathcal{M}^{o}_{l|r} = 1 - \exp\left(-\eta\, |d_{l} + d_{l|r}|\right)$, where $d_l$ is the left disparity and $d_{l|r}$ is the disparity of the right image projected to the left view.

\PAR{Stereo-Mono Fusion Loss.}
The mono-stereo fusion loss $\mathcal{L}_{ms}$ guides the fusion network at depth discontinuities with the occlusion mask to obtain a fused depth map, $\tilde{z}_f = \mathcal{M}^{o}_{l|r} z^m + (1 - \mathcal{M}^{o}_{l|r}) z^s$, using the following loss, 
\begin{equation}
\begin{aligned}
\mathcal{L}_{ms} &= \|z_f - \tilde{z}_f\|_1 .
\label{eq:pseudo-fusion}
\end{aligned}
\end{equation}

\vspace{-5pt}
\PAR{Ambient Image Consistency.}
The ambient luminance in a scene can vary by 14 orders of magnitude, inside a dark tunnel with bright sun at a tunnel exit, all in the same scene \cite{Myszkowski2016}. To tackle this extreme dynamic range, we reconstruct the ambient $\Lambda^{k_0}$ in the scene from the short exposure slice $\mu_k$, and sample $\Lambda^{HDR}$ from the HDR passive captures $I^4, I^5$. Then, novel scene images $\hat{I}_v^k$ can be expressed as,
\begin{align}
    \Lambda^{HDR}_v & = \mu_s(I^4_v+I^5_v-D_v^4-D_v^5)/(\mu_4+\mu_5),\\
    \Lambda^{k_0}_v & = \mu_k(I^4_v+I^5_v-D_v^4-D_v^5)/(\mu_4+\mu_5), \\
    \hat{I}_v^k & = \text{clip}\left(I_v^k - \Lambda^{k_0}_v + \Lambda^{HDR}_v,0,2^{10}\right),
\end{align}
with $\mu_s$ uniformly sampled in the interval from $[0.5\mu_k, 1.5\mu_k]$. We supervise the network by enforcing the depth to be consistent across different ambient illumination levels. 

\begin{figure*}[!ht]
    \vspace{-5mm}
    \centering
    \includegraphics[width=0.98\textwidth]{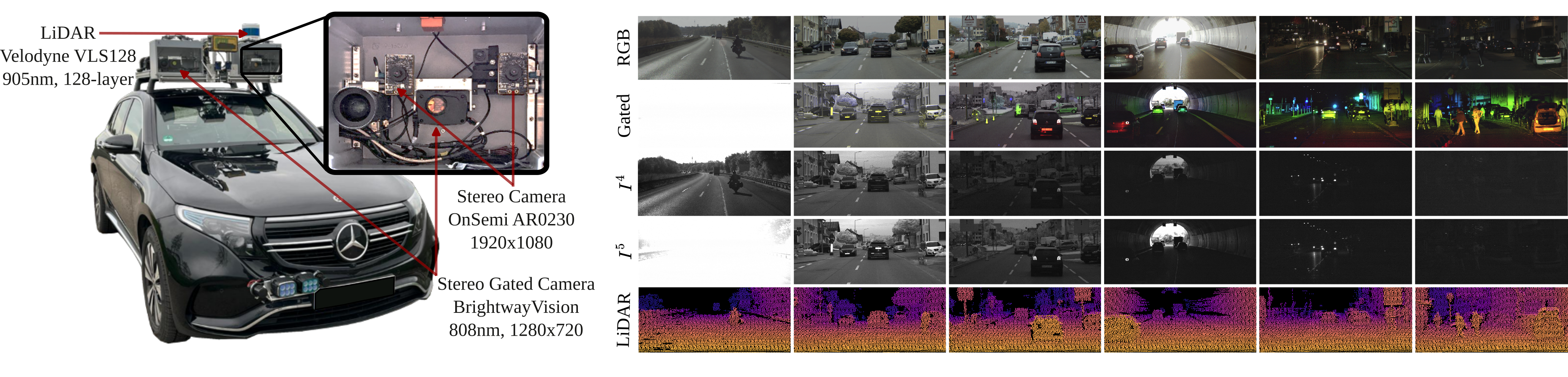}
    \vspace{-3mm}
    \caption{Illustration of the used sensor setup (left) and example captures from the wide-base gated stereo dataset (right). From top to bottom: RGB, Gated with red for slice 1, green for slice 2 and blue for slice 3, Gated Passive with low exposure time $I^4$, Gated Passive with high exposure time $I^5$, LiDAR. Note, the availability of a large number of frames with $\alpha C_k<I^k$.}
    \label{fig:dataset}
    \vspace{-4mm}
\end{figure*}

\PAR{Gated Reconstruction Loss.}
We adopt the cyclic gated reconstruction loss from~\cite{gated2gated}, which uses measured range intensity profiles $C_k(z)$ to reconstruct the input gated images from the predicted depth $z$, the albedo $\tilde{\alpha}$ and the ambient $\tilde{\Lambda}$. We estimate the $\tilde{\alpha}$ and $\tilde{\Lambda}$ from the context encoder through an additional U-Net like decoder, see Figure~\ref{fig:arch} and Supplemental Material. Specifically, the consistency loss models a gated slice as,
\begin{equation}
\begin{aligned}
\tilde{I}^k(z)\;&=\;\tilde{\alpha}\,C_k(z) + \tilde{\Lambda}.
\label{eq:gated_reconstruction}
\end{aligned}
\end{equation}
The loss term is based on the per-pixel difference and structural similarity as follows,
\begin{equation}
\begin{aligned}
\mathcal{L}_{recon} = \mathcal{L}_{p}(M_{g} \odot \tilde{I}^k(z), M_{g} \odot I^k)+\mathcal{L}_p(\tilde{\Lambda},\Lambda^{k_0}).
\label{eq:gated_reconstruction_loss}
\end{aligned}
\end{equation}
Similar to \cite{gated2gated} we utilize per-pixel SNR to obtain the gated consistency mask $M_g$.
See the Supplemental Material for a detailed derivation.
This loss enforces that the predicted depth is consistent with the simulated gated measurements. 

\PAR{Illuminator View Consistency.}
In the proposed gated stereo setup, we can enforce an additional depth consistency from the illuminator field of view. In this virtual camera view no shadows are visible as illustrated in Figure~\ref{fig:shadowloss}. This effectively makes the regions that are visible to the two cameras and the illuminator consistent. We use the gated consistency mask $M_{g}$ to supervise only regions that are illuminated by the laser and project the gated views $I_{l,r}$ into the laser field of view $I_{il|r,l}$, resulting in the loss,
\begin{equation}
\begin{aligned}
\mathcal{L}_{illum} &= \mathcal{L}_p(M_{g} \odot I_{il|l}, M_{g} \odot I_{il|r}) .
\label{eq:illum_consistency}
\end{aligned}
\end{equation}
\vspace{-17pt}
\PAR{Image Guided Depth Regularization.}
Following binocular and multi-view stereo methods \cite{Godard2017,Zhou2017}, we add an edge-aware smoothness loss $\mathcal{L}_{smooth}$ as regularization to the mean normalized inverse depth estimates $d$,
\begin{equation}
\begin{aligned}
\mathcal{L}_{smooth} &= |\nabla_x d|e^{-|\nabla_x I|} + |\nabla_y d|e^{-|\nabla_y I|}.
\label{eq:smoothness}
\end{aligned}
\end{equation}
\vspace{-17pt}
\PAR{Sparse LiDAR Supervision.}
The proposed gated stereo system has a higher update rate (\unit[24]{Hz}) than typical scanning LiDAR (\unit[10]{Hz}). Therefore, sparse LiDAR supervision can only be applied to samples fully in sync while all the previously presented self-supervised losses are applied to all samples. 
The LiDAR returns are first compensated for ego-motion, and then projected onto the image space. The supervision loss $\mathcal{L}_{sup}$ for view $v$ is,
\begin{equation}
\begin{aligned}
\mathcal{L}_{sup} &= \mathcal{M}_{v|s} \odot \|z_v - z_{v|s}^{*}\|_1,
\label{eq:lidar_supervision}
\end{aligned}
\end{equation}
where $\mathcal{M}_{v|s}$ is a binary mask indicating the projection of LiDAR points on the image, and $z_{v|s}^*$ is the ground-truth depth from a single LiDAR scan projected into the image $v$.

\vspace{-13pt}
\paragraph{Overall Training Loss.}

Combining all self-supervised and supervised loss components from above, we arrive at the following loss terms,
\vspace{-1mm}
\begin{align}
\mathcal{L}_{mono} &= c_1 \mathcal{L}_{recon} + c_2 \mathcal{L}_{sup} + c_3 \mathcal{L}_{smooth},
\label{eq:total_mono_loss}
\\
\mathcal{L}_{stereo} &= c_4 \mathcal{L}_{reproj} + c_5 \mathcal{L}_{recon} + c_6 \mathcal{L}_{illum} \nonumber \\
 &  + c_7 \mathcal{L}_{sup} + c_8 \mathcal{L}_{smooth},
\label{eq:total_stereo_loss}
\\
\mathcal{L}_{fusion} &= c_9 \mathcal{L}_{ms} + c_{10} \mathcal{L}_{sup} + c_{11} \mathcal{L}_{smooth},
\label{eq:total_fusion_loss}
\end{align}
\vspace{-1mm}
which we combine with scalar weights $c_{1, \ldots, 11}$ provided in the Supplemental Material.

\subsection{Implementation Details}
We first independently optimize the monocular and stereo networks using the losses presented in Sec.~\ref{sec:consistencylosses}. 
Both the stereo and monocular networks are trained using the same protocol using ADAMW \cite{ADAMW} with $\beta_1\,=\,0.9$, $\beta_2\,=\,0.999$, learning rate of $10^{-4}$ and of weight decay $10^{-2}$.
Finally, the fusion network is trained for 5 epochs using ADAMW and the losses described in Eq. \ref{eq:total_fusion_loss} with a learning rate of $3\cdot10^{-4}$. We used $\eta\,=\,0.05$ for generating occlusion masks referred in 
Equation \ref{eq:reproj_loss}.
For gated consistency masks, we set $\gamma=0.98,\ \theta=0.04$. 
All models are trained with input/output resolution of $1024 \times 512$.

\begin{table}[!t]
    \vspace{-0.2eM}
    \footnotesize
    \setlength{\tabcolsep}{4pt} 
    \setlength\extrarowheight{2pt}
    \centering
    \resizebox{.99\linewidth}{!}{
    \begin{tabular}{@{}c|lcccccccc@{}}
            \toprule
            & \multirow{2}{*}{\textbf{\textsc{Method}}} & \textbf{Modality} & \textbf{Train} & \textbf{RMSE}     & \textbf{ARD}   & \textbf{MAE}  & $\boldsymbol{\delta_1}$ & $\boldsymbol{\delta_2}$ & $\boldsymbol{\delta_3}$  \\ 
			&& &  & $\left[ m \right]$  &  & $\left[ m \right]$ & $\left[ \% \right]$ & $\left[ \% \right]$ & $\left[ \% \right]$\\
			\midrule
			\multicolumn{10}{c}{\textbf{Test Data -- Night (Evaluated on LiDAR Ground-Truth Points)}} \\
			\midrule
			\multirow{19}{*}{\rotatebox[origin=l]{90}{\parbox[c]{6.5cm}{\centering \textbf{\textsc{Comparison to state-of-the-art}}}}} 
			 & \textsc{Gated2Depth} \cite{gated2depth2019} & Mono-Gated & D & 16.15  &  0.17  &  8.07  &  75.70  &  92.74  &  96.47  \\ 
 & \textsc{Gated2Gated}  \cite{gated2gated} & Mono-Gated & MG & 14.08  &  0.19  &  7.95  &  79.84  &  92.95  &  96.59  \\ 
& \textsc{Sparse2Dense}  \cite{ma2018sparse} & Mono-Sparse & D & 9.97 &  0.11  &  5.22  &  87.06  &  95.77  &  98.20  \\ 
 & \textsc{KBNet}  \cite{wong2021unsupervised} & Mono-Sparse & D & 13.52  &  0.16  &  8.56  &  81.41  & \textbf{ 99.33 } & \textbf{ 99.66 } \\ 
 & \textsc{NLSPN} \cite{park2020nonRGBLidar} & Mono-Sparse & D & 12.19  &  0.09  &  5.42  &  89.63  &  96.84  &  99.03  \\ 
 & \textsc{PENet} \cite{hu2020PENetRGBLidar} & Mono-Sparse & D & 7.81  &  0.09  & \underline{ 3.59 } & \underline{ 93.68 } &  97.90  &  99.16  \\ 
 & \textsc{GuideNet} \cite{guidenetRGBLidar} & Mono-Sparse & D & \underline{7.50 } &  0.09  &  3.63  &  92.70  &  98.16  & \underline{ 99.35 } \\  
 & \textsc{PackNet} \cite{guizilini20203d} & Mono-RGB & M & 17.82  &  0.20  &  10.21  &  66.35  &  87.85  &  95.61  \\ 
 & \textsc{Monodepth2} \cite{godard2019digging} & Mono-RGB & M & 18.44  &  0.18  &  9.47  &  75.70  &  90.46  &  95.68  \\ 
 & \textsc{SimIPU} \cite{li2022simipu} & Mono-RGB & D & 15.78  &  0.18  &  8.71  &  76.25  &  90.84  &  96.44  \\ 
 & \textsc{AdaBins} \cite{bhat2021adabins} & Mono-RGB & D & 14.45  &  0.15  &  7.58  &  81.47  &  93.75  &  97.39  \\ 
 & \textsc{DPT} \cite{ranftl2021vision} & Mono-RGB & D & 12.15  &  0.12  &  6.31  &  85.38  &  95.94  &  98.42  \\ 
 & \textsc{DepthFormer} \cite{li2022depthformer} & Mono-RGB & D & 12.15  &  0.11  &  6.20  &  85.18  &  95.76  &  98.47  \\ 
 & \textsc{PSMNet} \cite{Chang2018} & Stereo-RGB & D & 27.98  &  0.27  &  16.02  &  50.77  &  74.77  &  85.93  \\ 
 & \textsc{STTR} \cite{li2021revisiting} & Stereo-RGB & D & 20.99  &  0.19  &  11.14  &  70.84  &  87.70  &  93.46  \\ 
 & \textsc{HSMNet} \cite{yang2019hsm} & Stereo-RGB & D & 12.42  &  0.09  &  5.87  &  88.41  &  96.08  &  98.50  \\ 
 & \textsc{ACVNet} \cite{xu2022attention} & Stereo-RGB & D & 11.70  & \underline{ 0.08 } &  5.25  &  89.91  &  96.33  &  98.47  \\ 
 & \textsc{RAFT-Stereo} \cite{lipson2021raft} & Stereo-RGB & D & 10.89  &  0.09  & 5.10 & 90.47 &  96.71  &  98.64  \\ 
 & \textbf{\textsc{Gated Stereo}} & Stereo-Gated & DGS & \textbf{6.39 } & \textbf{ 0.05 } & \textbf{ 2.25 } & \textbf{ 96.40 } & \underline{ 98.44 } & 99.24

\\
			\midrule
			\multicolumn{10}{c}{\textbf{Test Data -- Day (Evaluated on LiDAR Ground-Truth Points)}} \\
			\midrule
			\multirow{19}{*}{\rotatebox[origin=l]{90}{\parbox[c]{6.5cm}{\centering \textbf{\textsc{Comparison to state-of-the-art}}}}}
             & \textsc{Gated2Depth} \cite{gated2depth2019} & Mono-Gated & D & 28.68  &  0.22  &  14.76  &  66.68  &  82.76  &  87.96  \\ 
 & \textsc{Gated2Gated}  \cite{gated2gated} & Mono-Gated & MG & 16.87  &  0.21  &  9.51  &  73.93  &  92.15  &  96.10  \\ 
 & \textsc{Sparse2Dense}  \cite{ma2018sparse} & Mono-Sparse & D & 10.05  &  0.11  &  4.77  &  88.06  &  96.57  &  98.63  \\ 
 & \textsc{KBNet}  \cite{wong2021unsupervised} & Mono-Sparse & D & 15.27  &  0.17  &  9.54  &  78.54  & \textbf{ 99.31 } & \textbf{ 99.63 } \\ 
 & \textsc{NLSPN} \cite{park2020nonRGBLidar} & Mono-Sparse & D & 11.78  &  0.08  &  4.99  &  91.41  &  97.70  & \underline{ 99.24 } \\ 
 & \textsc{PENet} \cite{hu2020PENetRGBLidar} & Mono-Sparse & D & 8.54  &  0.09  &  3.82  &  93.78  &  97.69  &  98.94  \\ 
 & \textsc{GuideNet} \cite{guidenetRGBLidar} & Mono-Sparse & D & \underline{8.03 } &  0.09  & \underline{ 3.70 } &  93.23  &  98.12  &  99.21  \\ 
 & \textsc{PackNet} \cite{guizilini20203d} & Mono-RGB & M & 17.69  &  0.21  &  9.77  &  72.12  &  90.65  &  96.51  \\ 
 & \textsc{Monodepth2} \cite{godard2019digging} & Mono-RGB & M & 20.78  &  0.22  &  10.06  &  79.05  &  90.66  &  94.69  \\ 
 & \textsc{SimIPU} \cite{li2022simipu} & Mono-RGB & D & 14.33  &  0.14  &  7.50  &  81.77  &  94.01  &  97.92  \\ 
 & \textsc{AdaBins} \cite{bhat2021adabins} & Mono-RGB & D & 12.76  &  0.12  &  6.53  &  86.15  &  95.77  &  98.41  \\ 
 & \textsc{DPT} \cite{ranftl2021vision} & Mono-RGB & D & 11.29  &  0.09  &  5.52  &  89.56  &  96.83  &  98.79  \\ 
 & \textsc{DepthFormer} \cite{li2022depthformer} & Mono-RGB & D & 10.59  &  0.09  &  5.06  &  90.65  &  97.46  &  99.02  \\ 
 & \textsc{PSMNet} \cite{Chang2018} & Stereo-RGB & D & 32.13  &  0.28  &  18.09  &  53.82  &  74.91  &  84.96  \\ 
 & \textsc{STTR} \cite{li2021revisiting} & Stereo-RGB & D & 16.77  &  0.16  &  8.99  &  78.44  &  93.53  &  98.01  \\ 
 & \textsc{HSMNet} \cite{yang2019hsm} & Stereo-RGB & D & 10.36  &  0.08  &  4.69  &  92.47  &  97.93  &  99.11  \\ 
 & \textsc{ACVNet} \cite{xu2022attention} & Stereo-RGB & D & 9.40  & \underline{ 0.07 } &  4.08  & \underline{ 94.61 } &  98.36  & 99.12 \\ 
 & \textsc{RAFT-Stereo} \cite{lipson2021raft} & Stereo-RGB & D & 9.40 & \underline{ 0.07 } & 4.07 &  93.76  &  98.15  &  99.09  \\ 
 & \textbf{\textsc{Gated Stereo}} & Stereo-Gated & DGS & \textbf{7.11 } & \textbf{ 0.05 } & \textbf{ 2.25 } & \textbf{ 96.87 } & \underline{ 98.46 } &  99.11  
\\
			\midrule
            \bottomrule
    \end{tabular}
    }
    \vspace*{-5pt}
    \caption{\label{tab:results_g2d}\small Comparison of our proposed framework and state-of-the-art methods on the Gated Stereo test dataset. We compare our model to supervised and unsupervised approaches. M refers to methods that use temporal data for training, S for stereo supervision, G for gated consistency  and D for depth supervision. *~marked method are scaled with LiDAR ground-truth. Best results in each
    category are in \textbf{bold} and second best are
    \underline{underlined}.
	\vspace*{-4mm}
}
\end{table}
\begin{table*}[!t]
    \vspace*{-5mm}
	\centering
	\vspace{-0.2eM}
	\includegraphics[width=0.99\textwidth]{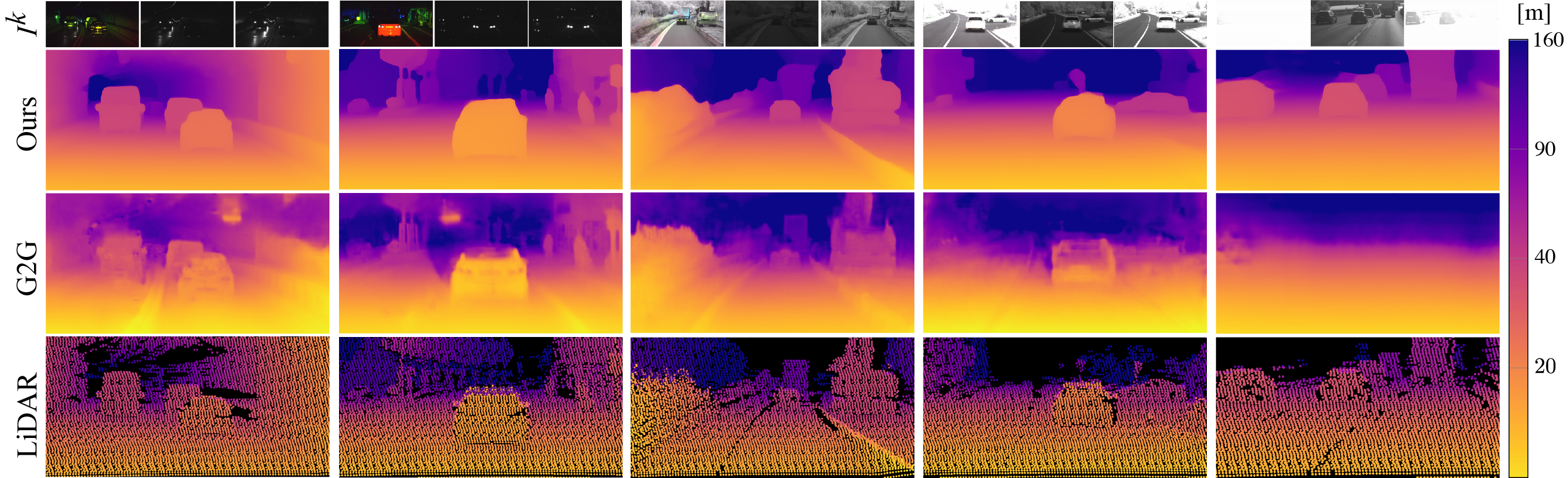}
	\vspace{-0.5eM}
	\captionof{figure}{The top row for each example shows the concatenated gated image $I^{1,2,3}$ and the corresponding passive images $I^4$ and $I^5$. The second row shows the depth map of our proposed method, the third row illustrates the results of Gated2Gated (G2G)\cite{gated2gated}, and the bottom row depicts the projected LiDAR point cloud into the gated view. Our method handles shadow areas and high-reflectivity targets much better than G2G. Furthermore, the HDR input allows to predict accurate depth even in bright conditions.}
	\label{fig:gs_vs_g2g}
	\vspace*{-0.4cm }
\end{table*}

\vspace{-5pt}
\section{Dataset}
\vspace{-1mm}
In this section, we describe the long-range depth dataset that we captured for training and testing. The dataset was acquired during a data collection campaign covering more than one thousand kilometers of driving in Southern Germany. We have equipped a testing vehicle with a long-range LiDAR system (Velodyne VLS128) with a range of up to \unit[200]{m}, an automotive RGB stereo camera (OnSemi AR0230 sensor) and a NIR gated stereo camera setup (BrightWayVision) with synchronization. The sensor setup is shown in Figure~\ref{fig:dataset} with all sensors mounted in a portable sensor cube, except for the LiDAR sensor. The RGB stereo camera has a resolution of 1920x1080 pixels and runs at 30 Hz capturing 12bit HDR images. The gated camera provides 10 bit images with a resolution of 1280x720 at a framerate of 120 Hz, which we split up into three slices plus two HDR-like additional ambient captures without active illumination. We use two vertical-cavity surface-emitting laser (VCSEL) modules as active illumination mounted on the front tow hitch. The lasers flood illuminate the scene at a peak power of \unit[500]{W} each, a wavelength of \unit[808]{nm} and laser pulse durations of \unit[240-370]{ns}. The maximum peak power is thereby limited due to eye-safety regulations. The mounted reference LiDAR system is running with 10 Hz and yields 128 lines. All sensors are calibrated and time-synchronized and Fig.~\ref{fig:dataset} provides visual examples. The dataset contains 107348 samples in day, nighttime, and varying weather conditions. After sub-selection for scenario diversity, we split the dataset into 54320 samples for training, 728 samples for validation and, 2463 samples for testing, see Supplemental Material for details.

\vspace{-1mm}
\section{Assessment}
In this section, we validate the proposed method experimentally. We investigate depth estimation at night, day and compared to existing depth estimation methods. Moreover, we validate design choices with ablation experiments. 
\PAR{Experimental Setup.}
We evaluate on the proposed test set consisting of 2463 (1269 day/1194 night) frames with high-resolution 128-layer LiDAR ground-truth measurements up to \unit[200]{m}. Unlike existing work \cite{gated2depth2019,gated2gated,Uhrig2017THREEDV} which was limited to \unit[80]{m}, we are therefore able to report results up to a distance of \unit[160]{m} to asses long-range depth prediction. 
Following \cite{eigen2014depth}, we evaluate depth using the metrics RMSE, MAE, ARD, and $\delta_i < 1.25i$ for $i \in {1,2,3}$ and split results for day and night. For fair comparison, \emph{all methods we compare to have been fine-tuned} on our dataset. Details on the fine-tuning of reference methods are given in Section~4.1. of the Supplemental Material.
\begin{table}[!t]
	\centering
	\setlength{\tabcolsep}{2pt}
	\resizebox{1.01\linewidth}{!}{
		\begin{tabular}{l|lccccccccc}
			\toprule
			& \multirow{2}{*}{\textbf{Modality}} &  \textbf{HDR} &   \textbf{Ambient} &  \textbf{Cycle} &  \textbf{Warp} &  \textbf{RMSE} &  \textbf{MAE}  & $\boldsymbol{\delta_1}$ & $\boldsymbol{\delta_2}$ & $\boldsymbol{\delta_3}$  \\ 
			& & & \textbf{Con.} & \textbf{Con.} & \textbf{Con.} & $\left[ m \right]$ & $\left[ m \right]$ & $\left[ \% \right]$ & $\left[ \% \right]$ & $\left[ \% \right]$\\
			\midrule
			\multicolumn{11}{c}{\textbf{Test Data -- Night (Evaluated on LiDAR Ground-Truth Points)}} \\
			\midrule
			\midrule
			\multirow{7}{*}{\rotatebox[origin=l]{90}{\parbox[c]{3cm}{\centering \textbf{\textsc{Ablation}}}}} 
			 & Mono-Gated & \xmark & \xmark & \xmark & \xmark & 8.03  &  3.36  &  93.45  &  97.56  &  98.91  \\ 
 & Mono-Gated & \cmark & \cmark & \cmark & \xmark & \underline{7.07 } &  2.60  & \underline{ 95.91 } & \underline{ 98.14 } & \underline{ 99.09 } \\ 
 & Stereo-Gated \cite{lipson2021raft}%
 & \xmark & \xmark & \xmark & \xmark & 7.92  &  3.06  &  95.23  &  97.98  &  99.00  \\ 
 & Stereo-Gated & \cmark & \cmark & \xmark & \xmark & 7.38  &  2.41  &  95.63  &  98.01  &  99.02  \\ 
 & Stereo-Gated & \cmark & \cmark & \cmark & \xmark & 7.72  &  2.55  &  95.31  &  97.86  &  98.89  \\ 
 & Stereo-Gated & \cmark & \cmark & \cmark & \cmark & 7.33  & \underline{ 2.39 } &  95.84  &  98.09  &  99.02  \\ 
 & \textbf{Mono+Stereo-Gated} & \cmark & \cmark & \cmark & \cmark & \textbf{6.39 } & \textbf{ 2.25 } & \textbf{ 96.40 } & \textbf{ 98.44 } & \textbf{ 99.24 }  

\\
			\midrule
			\multicolumn{11}{c}{\textbf{Test Data -- Day (Evaluated on LiDAR Ground-Truth Points)}} \\
			\midrule
			\multirow{7}{*}{\rotatebox[origin=l]{90}{\parbox[c]{3cm}{\centering \textbf{\textsc{Ablation}}}}}
             & Mono-Gated & \xmark & \xmark & \xmark & \xmark & 11.93  &  5.31  &  90.15  &  95.62  &  97.73  \\ 
 & Mono-Gated & \cmark & \cmark & \cmark & \xmark & 9.26  &  3.66  &  94.69  &  97.84  &  98.88  \\ 
 & Stereo-Gated \cite{lipson2021raft}%
 & \xmark & \xmark & \xmark & \xmark & 9.77  &  4.03  &  92.15  &  96.69  &  98.28  \\ 
 & Stereo-Gated & \cmark & \cmark & \xmark & \xmark & 7.63  &  2.31  &  96.42  &  98.18  &  98.98  \\ 
 & Stereo-Gated & \cmark & \cmark & \cmark & \xmark & 7.87  &  2.27  &  96.46  &  98.13  &  98.92  \\ 
 & Stereo-Gated & \cmark & \cmark & \cmark & \cmark & \underline{7.47 } & \textbf{ 2.15 } & \underline{ 96.72 } & \underline{ 98.29 } & \underline{ 99.00 } \\ 
 & \textbf{Mono+Stereo-Gated} & \cmark & \cmark & \cmark & \cmark & \textbf{7.11 } & \underline{ 2.25 } & \textbf{ 96.87 } & \textbf{ 98.46 } & \textbf{ 99.11 }  

\\
			\midrule
            \bottomrule

	\end{tabular}}
	\vspace*{-5pt}
	\caption{Ablation studies evaluated on the proposed \textbf{Gated Stereo} test dataset. We investigate different input modalities, feature encoders, and loss combinations for the monocular and stereo network. Our final fusion model outperforms all other methods by a significant margin.}
	\label{tab:ablation}
	\vspace{-4mm}
\end{table}

\begin{figure*}[!t]
\vspace*{-1eM}
	\centering
	\begin{minipage}[t]{0.49\linewidth}
		\scriptsize
		\subcaption{\scriptsize Night: \textbf{Gated Stereo} is able to predict fine grained details even for far distances.  }
		\vspace{0.05cm}
		\renewcommand{\arraystretch}{0.7}
\setlength{\tabcolsep}{2pt}
\begin{tabular}{ccc}
	RGB & 
	Gated &
	LiDAR \\
	
	\includegraphics[width=0.31\columnwidth]{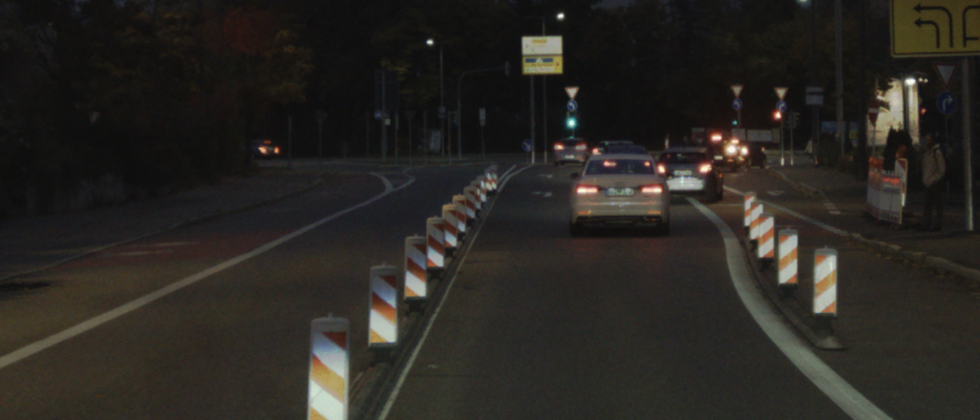} &
	\includegraphics[width=0.31\columnwidth]{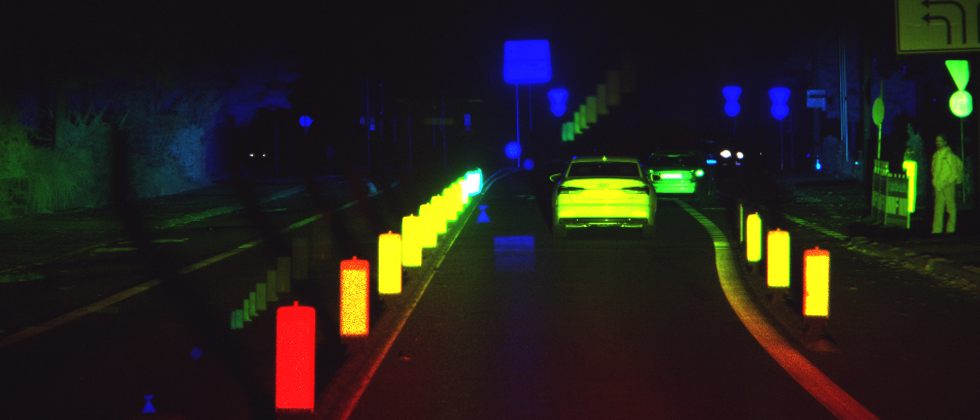} &
	\includegraphics[width=0.31\columnwidth]{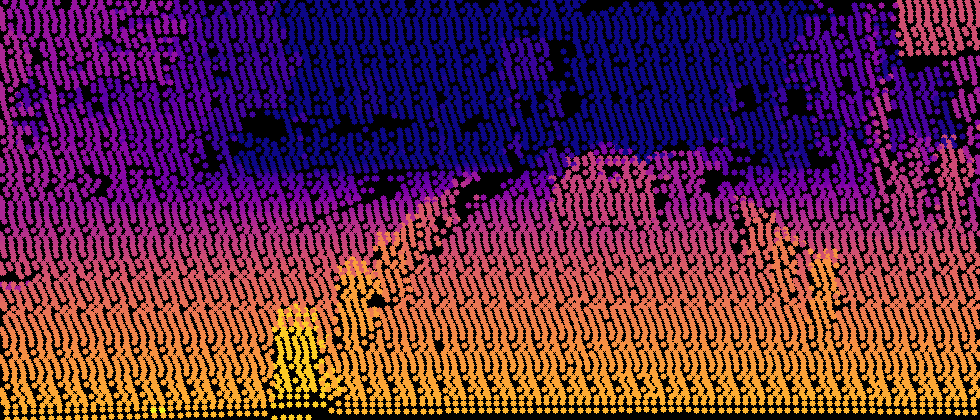} 
	\\
	
	\textbf{Gated Stereo} &
	Gated2Gated \cite{gated2gated} &
    Sparse2Dense  \cite{ma2018sparse} 
    \\

	\includegraphics[width=0.31\columnwidth]{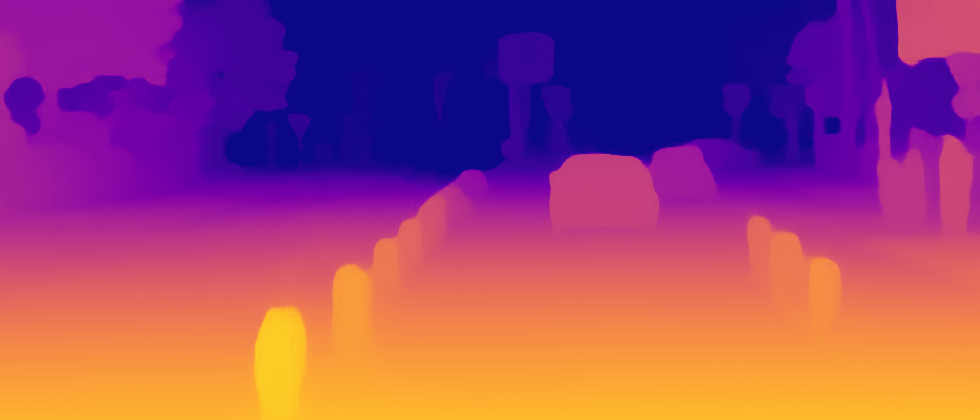} &
	\includegraphics[width=0.31\columnwidth]{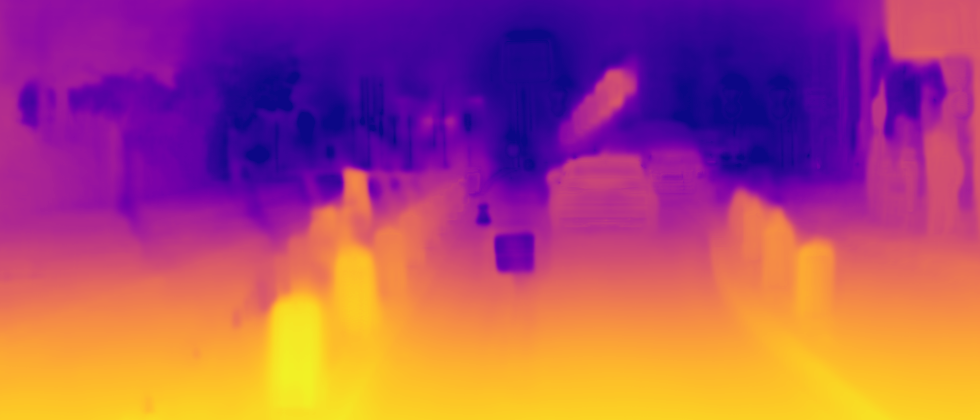} &
	\includegraphics[width=0.31\columnwidth]{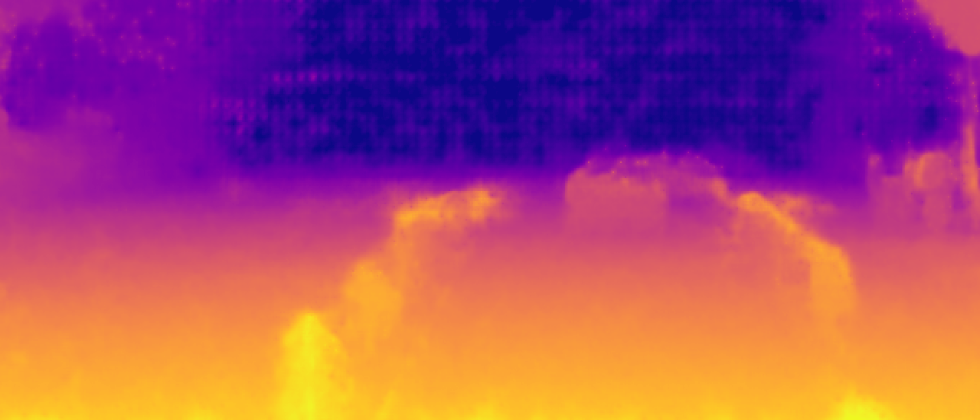} 
	\\

	Monodepth2 \cite{godard2019digging} &
	Depthformer \cite{li2022depthformer} & 
    Raft-Stereo \cite{lipson2021raft}
	\\
	
	\includegraphics[width=0.31\columnwidth]{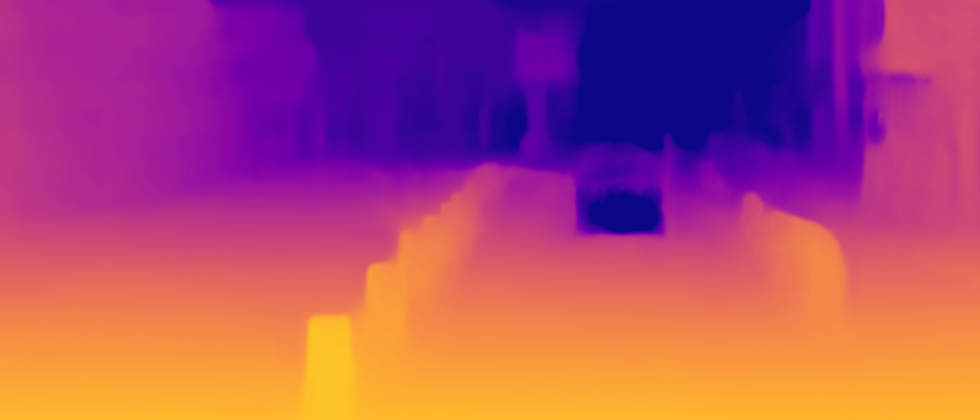} &
	\includegraphics[width=0.31\columnwidth]{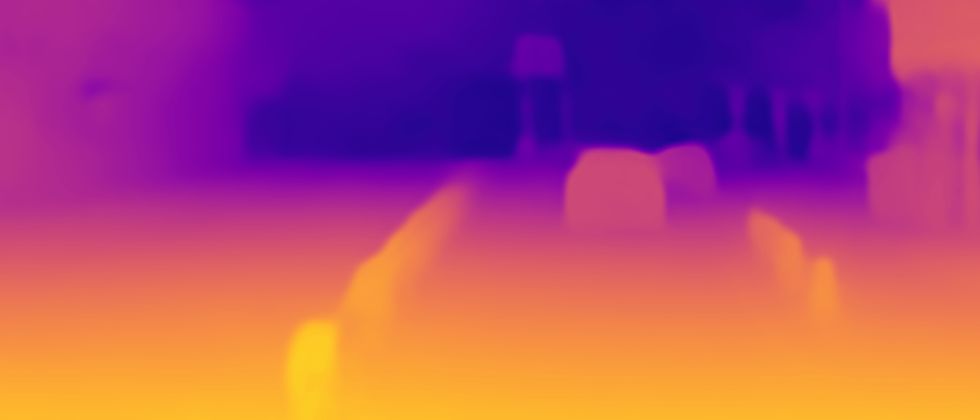} &
	\includegraphics[width=0.31\columnwidth]{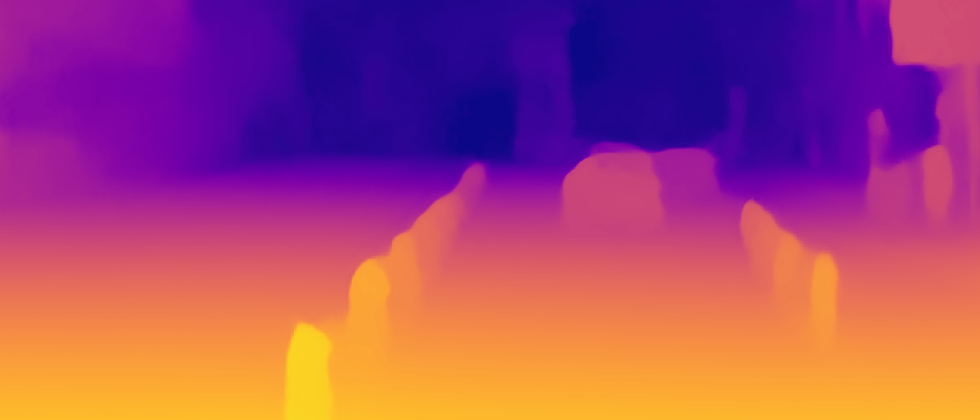} 
	\\
\end{tabular}

		\label{fig:comp_ref_methods_night}
	\end{minipage}%
	\centering
	\begin{minipage}[t]{0.49\linewidth}
		\scriptsize
		\subcaption{\scriptsize Day: \textbf{Gated Stereo} is able to handle bright sunlight conditions.}
		\vspace{0.05cm}
		\renewcommand{\arraystretch}{0.7}
\setlength{\tabcolsep}{2pt}
\begin{tabular}{@{}ccc@{}}
	RGB & 
	Gated &
	LiDAR \\
	
	\includegraphics[width=0.31\columnwidth]{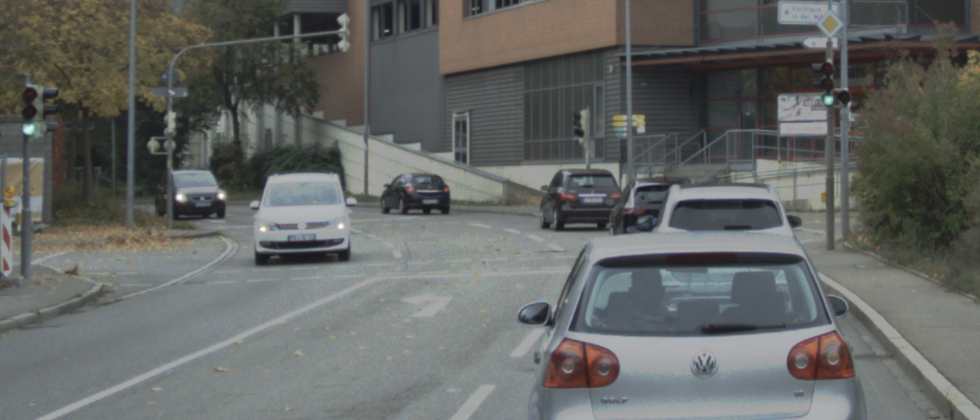} &
	\includegraphics[width=0.31\columnwidth]{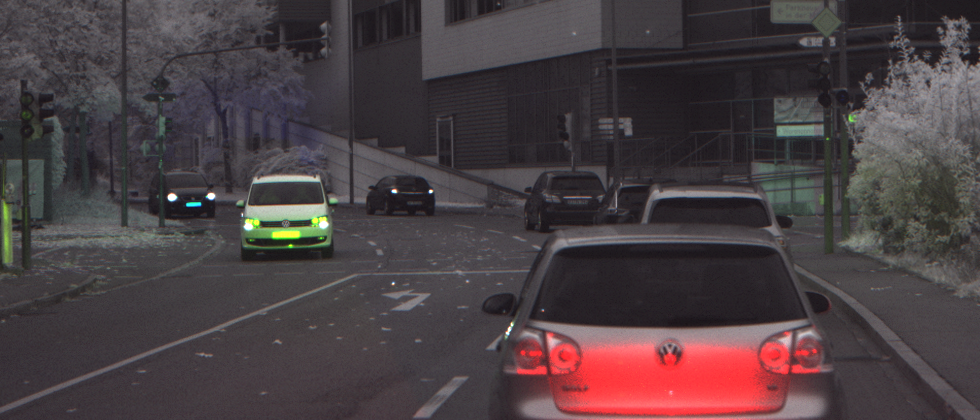} &
	\includegraphics[width=0.31\columnwidth]{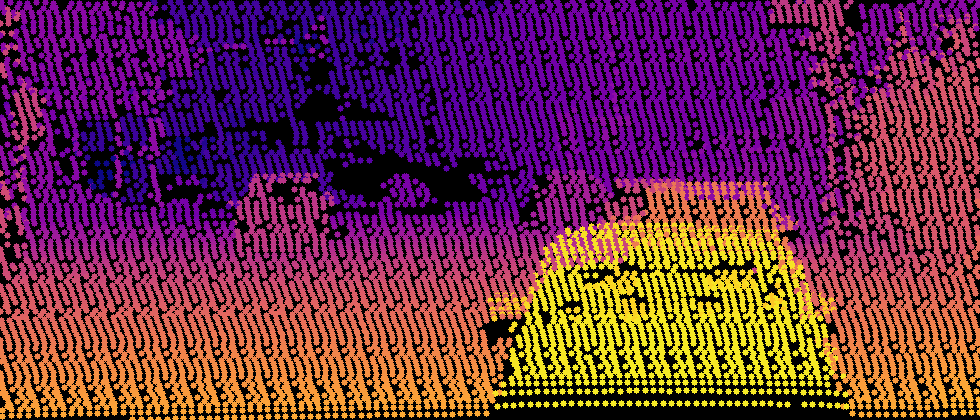} \\

    \textbf{Gated Stereo} &
	Gated2Gated \cite{gated2gated} &
    Sparse2Dense  \cite{ma2018sparse} 
    \\
    
	\includegraphics[width=0.31\columnwidth]{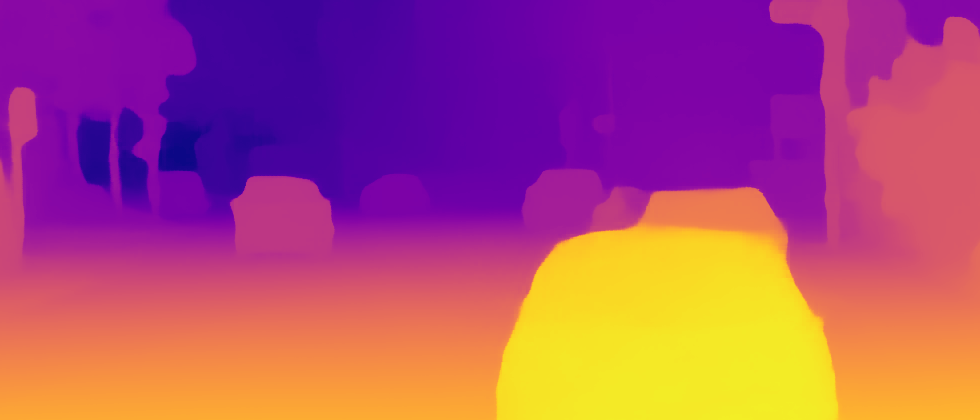} &
	\includegraphics[width=0.31\columnwidth]{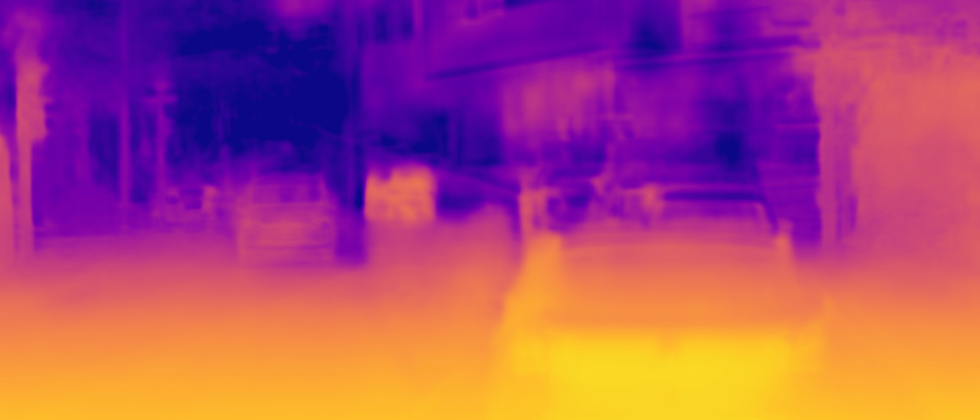} &
	\includegraphics[width=0.31\columnwidth]{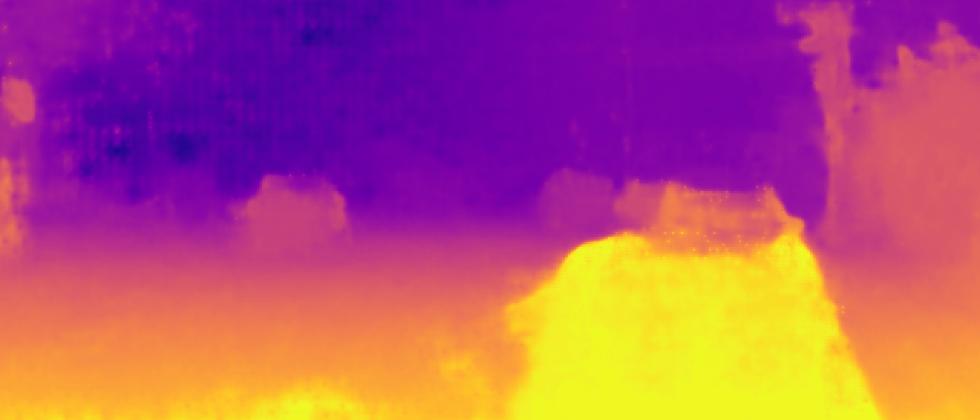} 
    \\
    
	Monodepth2 \cite{godard2019digging} &
	Depthformer \cite{li2022depthformer} & 
    Raft-Stereo \cite{lipson2021raft}
	\\
	
	\includegraphics[width=0.31\columnwidth]{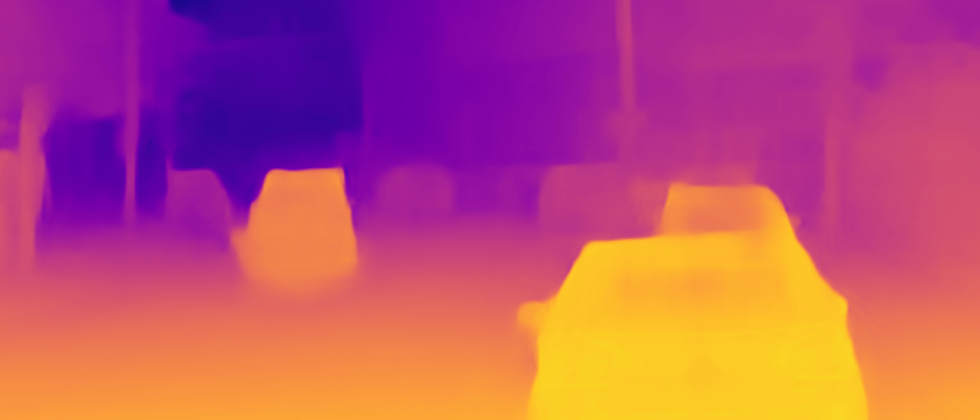} &
	\includegraphics[width=0.31\columnwidth]{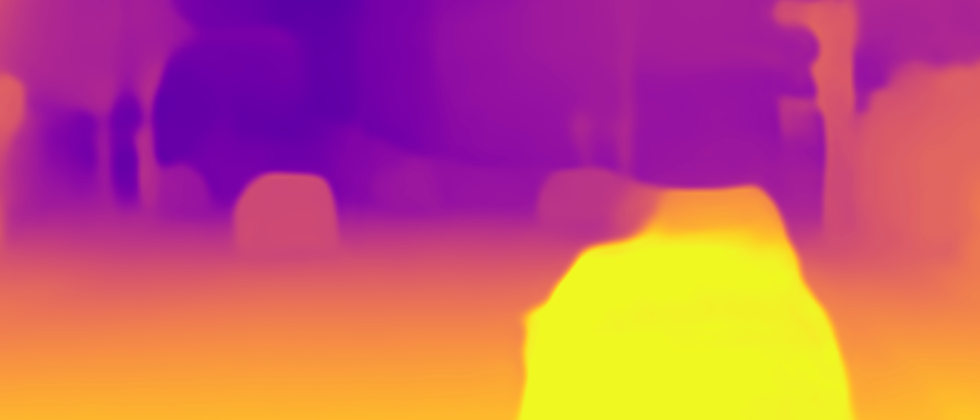} &
	\includegraphics[width=0.31\columnwidth]{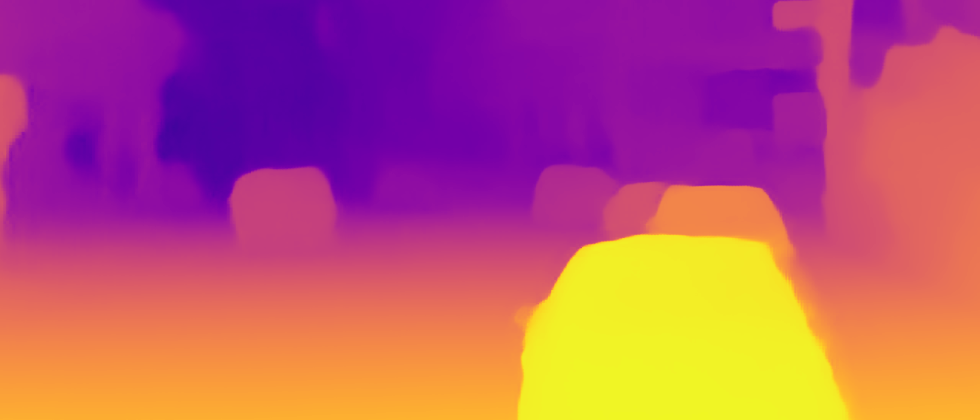} 
	\\
\end{tabular}
		\label{fig:comp_ref_methods_day}
	\end{minipage}%
	\centering
	\begin{minipage}[t]{0.02\linewidth}
		\scriptsize
		\begin{tabular}{@{}>{\centering\arraybackslash}m{0.2cm}}
			\multirow{1}{*}[-0.74cm]{\hspace{-0.042cm}[m]} \\
			\multirow{3}{*}[-0.72cm]{\hspace{-0.2cm}\includegraphics[height=3.95cm]{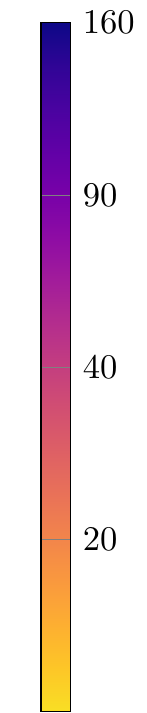}}\\
			
		\end{tabular}
	\end{minipage}
	\begin{minipage}[t]{0.49\linewidth}
		\scriptsize
		\vspace{-0.30cm}
		\renewcommand{\arraystretch}{0.7}
\setlength{\tabcolsep}{2pt}
\begin{tabular}{ccc}
	RGB & 
	Gated &
	LiDAR \\
	
	\includegraphics[width=0.31\columnwidth]{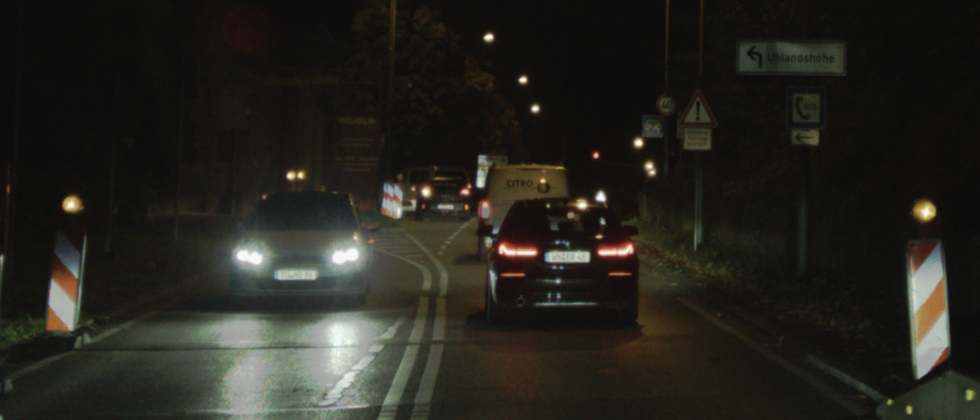} &
	\includegraphics[width=0.31\columnwidth]{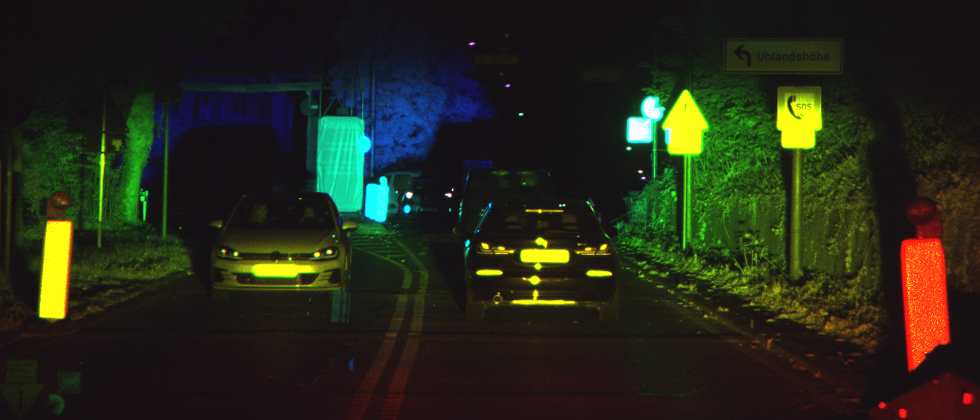} &
	\includegraphics[width=0.31\columnwidth]{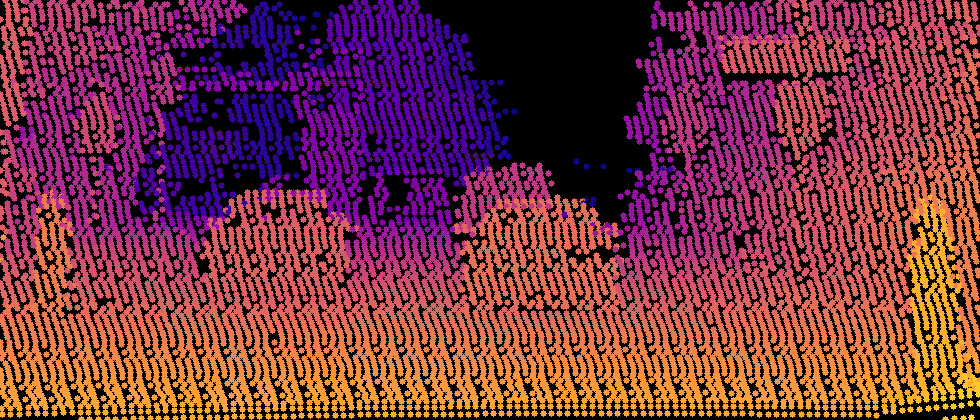} 
	\\
	
	\textbf{Gated Stereo} &
	Gated2Gated \cite{gated2gated} &
    Sparse2Dense  \cite{ma2018sparse} 
    \\

	\includegraphics[width=0.31\columnwidth]{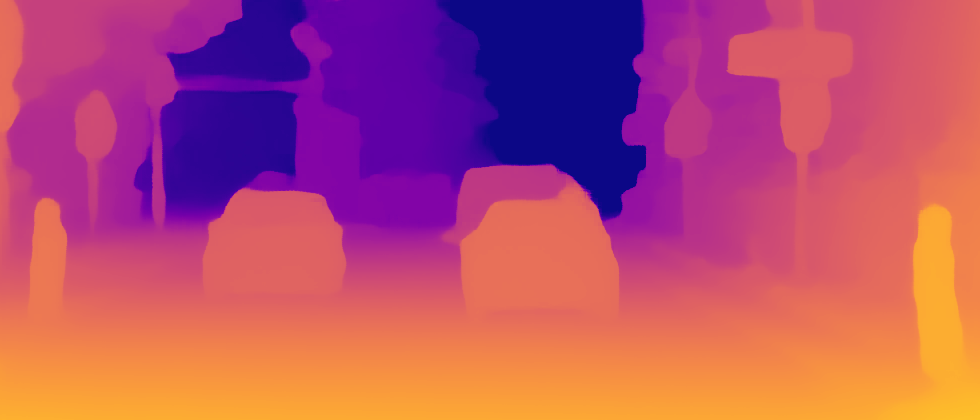} &
	\includegraphics[width=0.31\columnwidth]{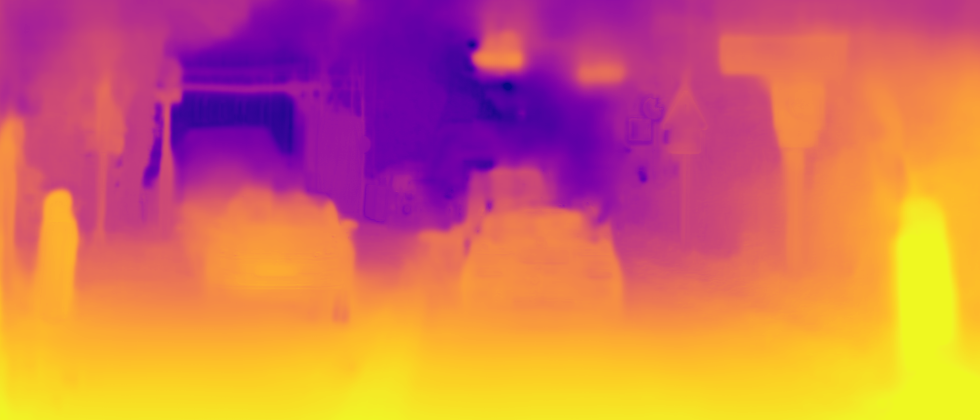} &
	\includegraphics[width=0.31\columnwidth]{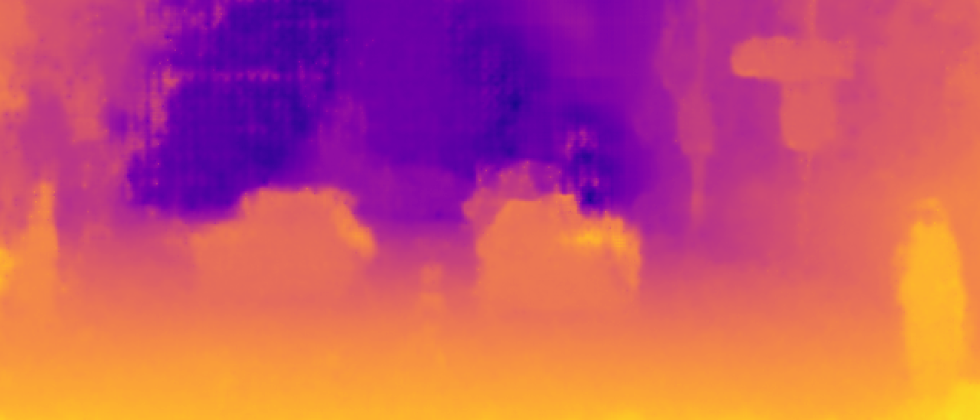} 
	\\

	Monodepth2 \cite{godard2019digging} &
	Depthformer \cite{li2022depthformer} & 
    Raft-Stereo \cite{lipson2021raft}
	\\
	
	\includegraphics[width=0.31\columnwidth]{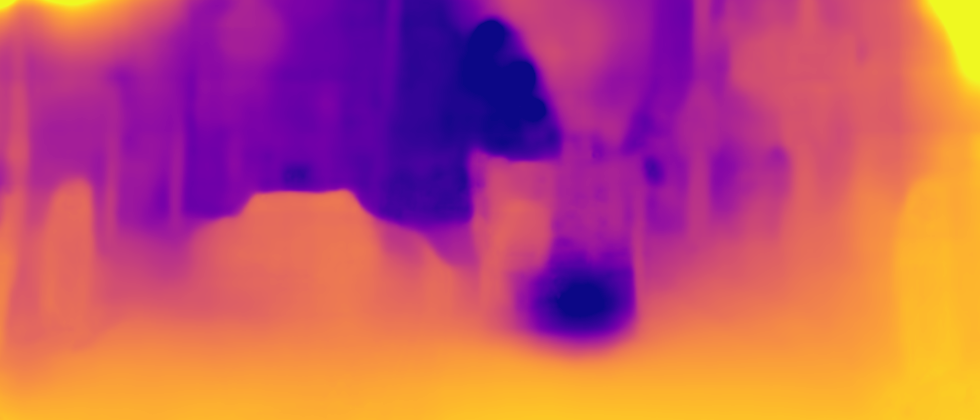} &
	\includegraphics[width=0.31\columnwidth]{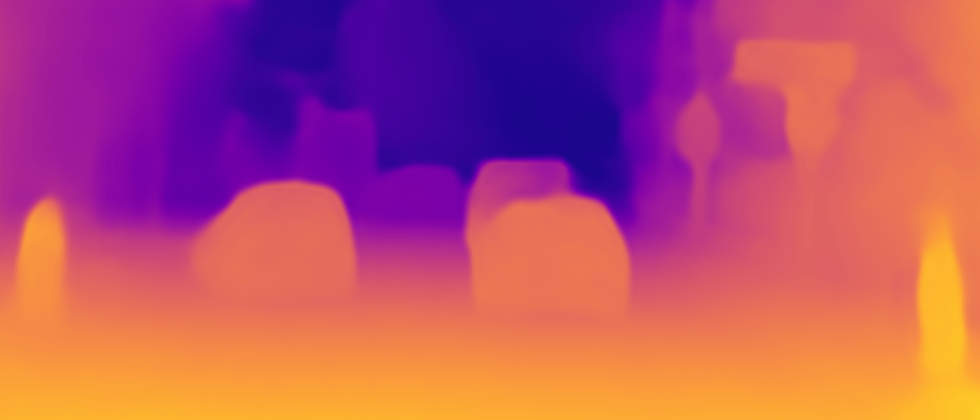} &
	\includegraphics[width=0.31\columnwidth]{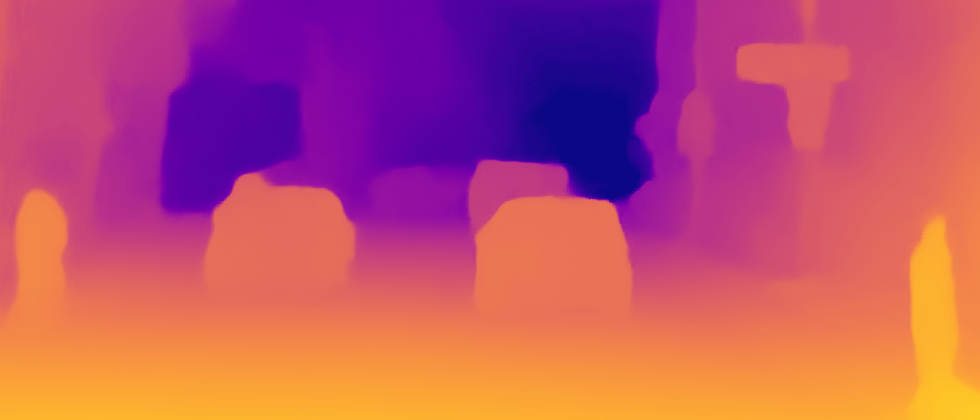} 
	\\
\end{tabular}

		\label{fig:comp_ref_methods_night}
	\end{minipage}%
	\centering
	\begin{minipage}[t]{0.49\linewidth}
		\scriptsize
		\vspace{-0.3cm}
		\renewcommand{\arraystretch}{0.7}
\setlength{\tabcolsep}{2pt}
\begin{tabular}{@{}ccc@{}}
	RGB & 
	Gated &
	LiDAR \\
	
	\includegraphics[width=0.31\columnwidth]{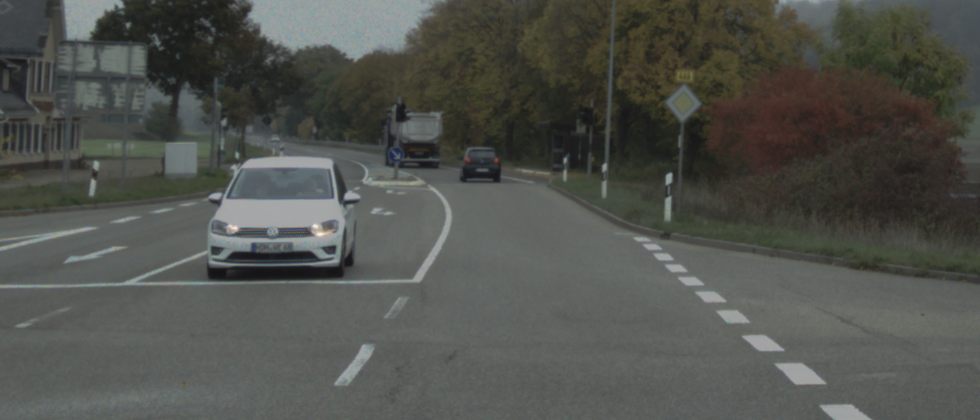} &
	\includegraphics[width=0.31\columnwidth]{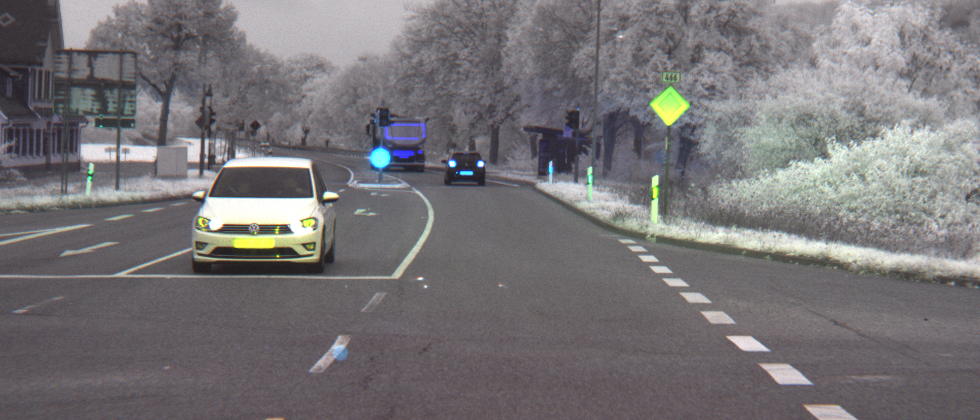} &
	\includegraphics[width=0.31\columnwidth]{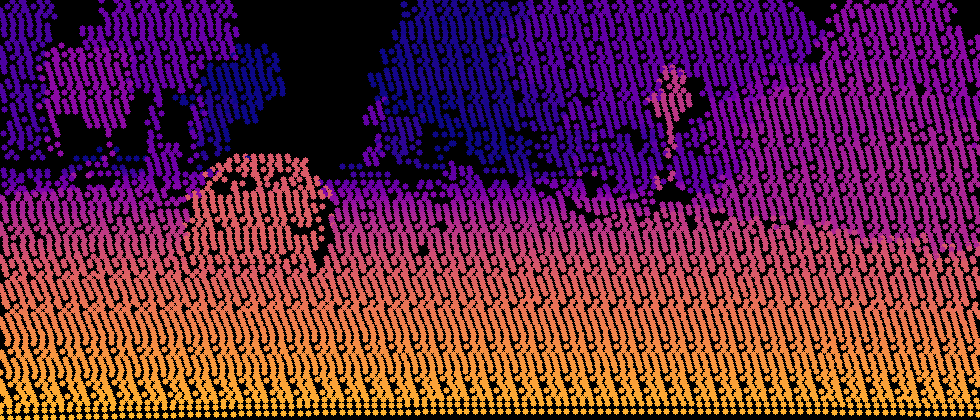} \\

    \textbf{Gated Stereo} &
	Gated2Gated \cite{gated2gated} &
    Sparse2Dense  \cite{ma2018sparse} 
    \\
    
	\includegraphics[width=0.31\columnwidth]{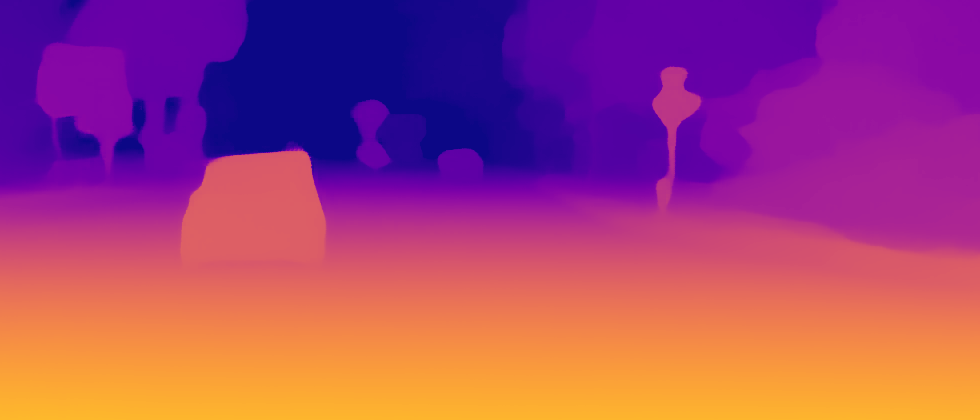} &
	\includegraphics[width=0.31\columnwidth]{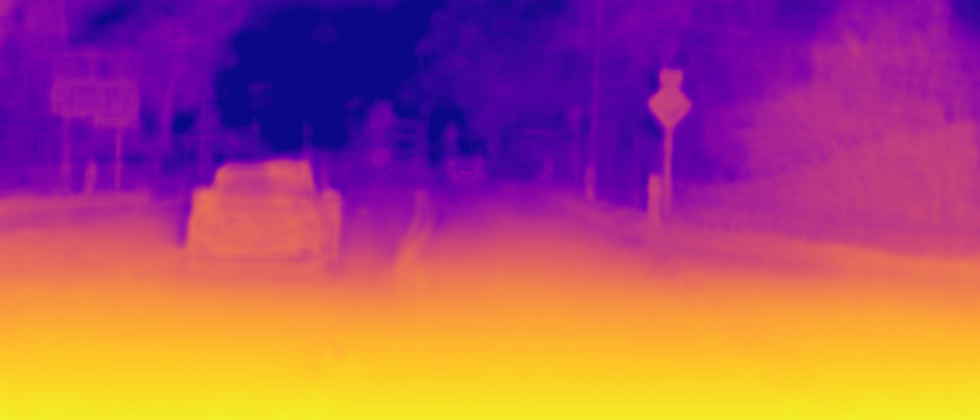} &
	\includegraphics[width=0.31\columnwidth]{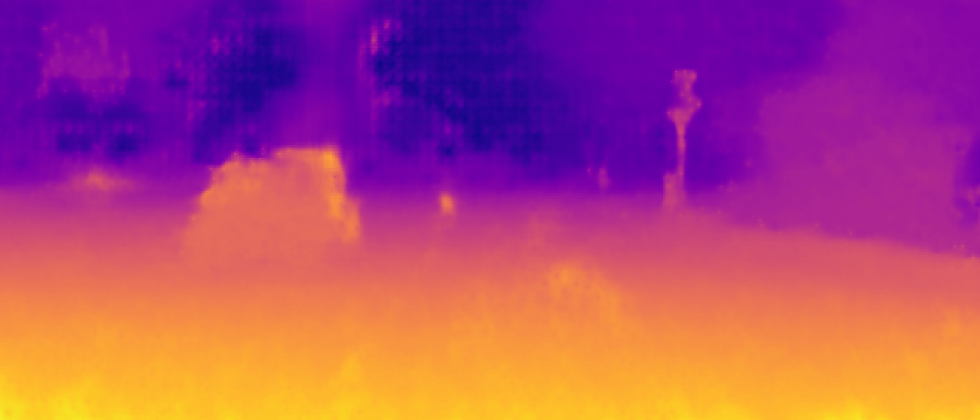} 
    \\
    
	Monodepth2 \cite{godard2019digging} &
	Depthformer \cite{li2022depthformer} & 
    Raft-Stereo \cite{lipson2021raft}
	\\
	
	\includegraphics[width=0.31\columnwidth]{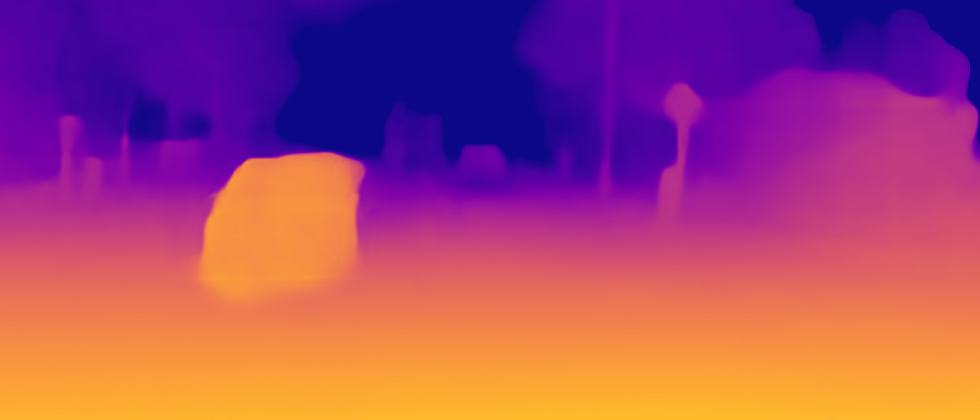} &
	\includegraphics[width=0.31\columnwidth]{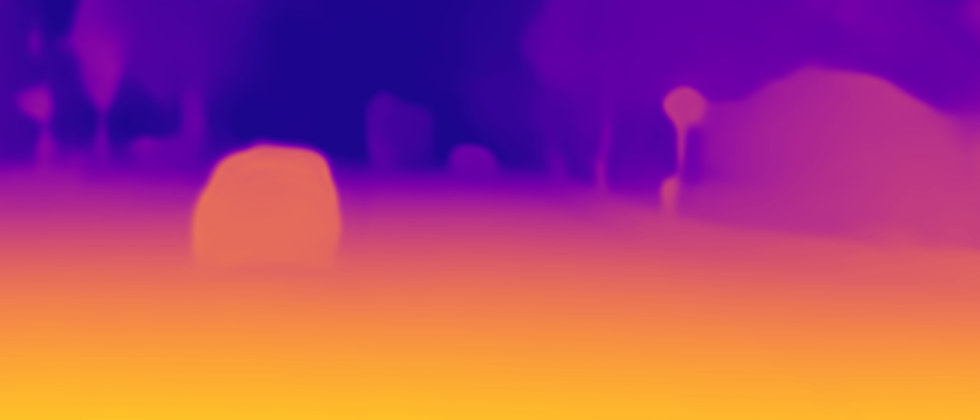} &
	\includegraphics[width=0.31\columnwidth]{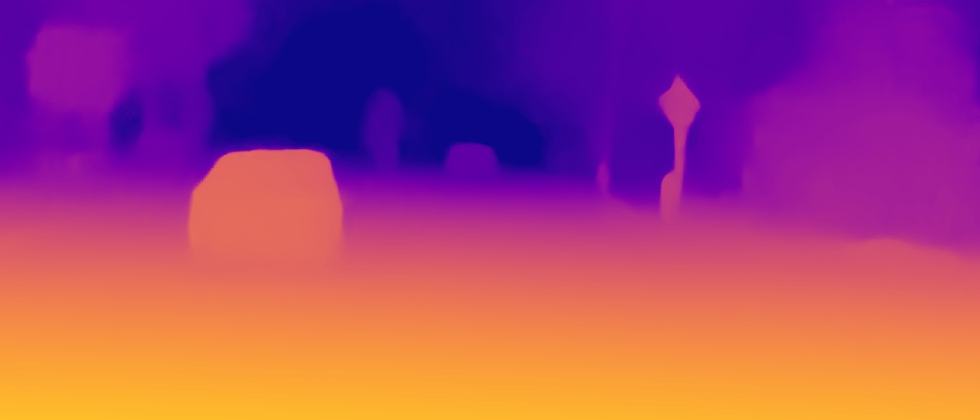} 
	\\
\end{tabular}
		\label{fig:comp_ref_methods_day}
	\end{minipage}%
	\centering
	\begin{minipage}[t]{0.02\linewidth}
		\scriptsize
		\begin{tabular}{@{}>{\centering\arraybackslash}m{0.2cm}}
			\multirow{1}{*}[-0.0cm]{\hspace{-0.042cm}[m]} \\
			\multirow{3}{*}[-0.0cm]{\hspace{-0.2cm}\includegraphics[height=3.95cm]{imgs/results/colorbar_depth_vertical.pdf}}\\
			
		\end{tabular}
	\end{minipage}%
	\vspace*{-3mm}
	\caption{\textbf{Qualitative comparison of \textbf{Gated Stereo} and existing methods}. For (a) night and (b) day conditions, Gated Stereo predicts sharper depth maps than existing methods. (In the gated image red refers to $I^1$, green to $I^2$, and blue to $I^3$).}
	\label{fig:comp_ref_methods}
	\vspace{-5mm}
\end{figure*}

\PAR{Depth Reconstruction.}
Qualitative results are presented in Figure~\ref{fig:comp_ref_methods} and quantitative results in Table~\ref{tab:results_g2d}. Here, we compare against two recent gated \cite{gated2depth2019,gated2gated}, six monocular RGB \cite{guizilini20203d,godard2019digging,li2022simipu,bhat2021adabins,ranftl2021vision,li2022depthformer}, five stereo RGB \cite{Chang2018,li2021revisiting,yang2019hsm,xu2022attention,lipson2021raft} and five monocular+LiDAR \cite{ma2018sparse,wong2021unsupervised, guidenetRGBLidar, park2020nonRGBLidar, hu2020PENetRGBLidar} methods. Comparing Gated Stereo to the next best stereo method RAFT-Stereo~\cite{lipson2021raft}, our method reduces error by \unit[45]{\%} and \unit[1.8]{m} in MAE in day conditions. In night conditions, the error is reduced by \unit[56]{\%} and \unit[2.9]{m} MAE. Qualitatively this improvement is visible in sharper edges and less washed-out depth estimates. Fine details, including thin poles, are better visible due to the structure-aware refinement achieved through the monocular depth outputs. 
The next best gated method, Gated2Gated \cite{gated2gated} achieves a \unit[9.51]{m} MAE in day conditions and \unit[7.95]{m} MAE in night conditions. Here, the performance drops significantly in day conditions due to strong ambient illumination, while Gated Stereo is capable of making use of the passive captures. This is also visible in the shown qualitative Figure~\ref{fig:gs_vs_g2g}, where Gated Stereo maintains high-quality depth outputs, while Gated2Gated fails.

Overall, we report a reduction of \unit[74]{\%} in MAE error compared to existing gated methods. Comparing to the best monocular RGB method, Depthformer \cite{li2022depthformer}, textures are often wrongly interpreted as rough surfaces missing smoothness. Lastly, we compare to monocular + LiDAR methods. Note, that the methods are fed with ground-truth points and therefore achieve competitive quantitative results on par with the best stereo methods. Qualitatively, the methods are not capable of interpolating plausible depth maps, which are instead washed out, and we find that problematic texture interpretation is carried over from monocular depth estimation methods.

\PAR{Ablation Experiments.}
To validate the contributions of each component of the proposed method, we report ablation experiments in Table~\ref{tab:ablation}, see Supplemental Material for qualitative results. In the following, we compare the MAE of the different models averaged over day and night. The starting point for our analysis is the monocular gated estimation using the proposed monocular branch with LiDAR supervision only. This method outperforms the best monocular RGB approach \cite{li2022depthformer} by \unit[23]{\%} lower MAE error. Next, the concatenated passive images and the active slices result in an added reduction of \unit[28]{\%} MAE error. We analyze RAFT-Stereo with stereo gated images and HDR-like passive frames as input. With additional Ambient Aware Consistency and the proposed backbone, we reduce the MAE error by \unit[25]{\%} compared to the next monocular gated approach and by \unit[36]{\%} to a native RAFT Stereo network with gated input. The HR-Former backbone alone contributes about \unit[10]{\%} of the \unit[33]{\%} reduction in MAE.
By adding the Gated Consistency loss and the warping losses for left-right consistency across views and illuminator the error further decreased by \unit[4]{\%}.
Finally, the fusion stage combining the monocular and stereo outputs preserves the fine structures from the monocular model and the long-range accuracy of the stereo model, results in an reduction of \unit[48]{\%} in MAE error when compared to monocular gating.

\vspace{-2mm}
\section{Conclusion}
\vspace{-1mm}
We present Gated Stereo, a long-range active multi-view depth estimation method. The proposed method predicts dense depth from synchronized gated stereo pairs acquired in a wide-baseline setup. 
The architecture comprises a stereo network and per-view monocular and stereo-mono fusion networks. All of these sub-networks utilize both active and passive images to extract depth cues. Stereo cues can be ambiguous, e.g., due to occlusion and repeated structure. Similarly, monocular gated cues can be insufficient in bright ambient illumination and at long range. To this end, our proposed approach predicts stereo \emph{and} per-camera monocular depth and finally fuses the two to obtain a single high-quality depth map. The different parts of the network are semi-supervised with sparse LiDAR supervision and a set of self-supervised losses that ensures consistency between different predicted outputs.
We train and validate the proposed method on a new long-range automotive dataset with a maximum depth range twice as long as prior work. The proposed method achieves \unit[50]{\%} better mean absolute depth error than the next best method on stereo RGB images and \unit[74]{\%} better than the next best existing gated method. 
In the future, we hope that the proposed method may allow us to solve novel 3D vision tasks that today's LiDAR systems cannot solve due to their angular resolution, such as detecting unseen small objects as lost debris at long distances and high-quality road edge and lane detection.

\vspace{0.2\baselineskip}\PAR{Acknowledgments} This work was supported in the AI-SEE project with funding from the FFG, BMBF, and NRC-IRA. We thank the Federal Ministry for Economic Affairs and Energy for support via the PEGASUS-family project {``VVM-Verification and Validation Methods for Automated Vehicles Level 4 and 5''}. Felix Heide was supported by an NSF CAREER Award (2047359), a Packard Foundation Fellowship, a Sloan Research Fellowship, a Sony Young Faculty Award, a Project X Innovation Award, and an Amazon Science Research Award.

{\small
\bibliographystyle{ieee_fullname}
\interlinepenalty=10000
\bibliography{refs}
}

\end{document}